\documentclass[runningheads]{llncs}

\usepackage{eccv}

\usepackage{eccvabbrv}

\usepackage{graphicx}
\usepackage{booktabs}

\usepackage[accsupp]{axessibility}  

\usepackage{hyperref}

\usepackage{orcidlink}

\usepackage{amsmath}
\usepackage{amssymb}
\usepackage{xcolor}
\usepackage{colortbl}
\usepackage{hhline}
\usepackage{bbold}
\usepackage{pifont}
\usepackage{multirow}
\usepackage{wrapfig}
\usepackage[ruled,linesnumbered]{algorithm2e}

\begin{document}

\title{Combining Discrepancy-Confusion Uncertainty and Calibration Diversity for Active Fine-Grained Image Classification} 

\titlerunning{DECERN}

\author{Yinghao Jin\orcidlink{0009-0001-1643-7712} \and
Xi Yang\thanks{Corresponding Author}\orcidlink{0000-0001-5039-3680}}

\authorrunning{Y.Jin et al.}

\institute{Jilin University, China}

\maketitle
\begin{abstract}
    Active learning (AL) aims to build high-quality labeled datasets by iteratively selecting the most informative samples from an unlabeled pool under limited annotation budgets.
    However, in fine-grained image classification, assessing this informativeness reliably is especially challenging due to subtle differences between classes.
    In this paper, we introduce a novel active learning method, 
    combining \underline{\textbf{D}}iscr\underline{\textbf{E}}pancy-\underline{\textbf{C}}onfusion unc\underline{\textbf{E}}rtainty and calib\underline{\textbf{R}}atio\underline{\textbf{N}} diversity for active fine-grained image classification (\textbf{DECERN}), 
    to effectively perceive the distinctiveness between fine-grained images and evaluate the sample value.
    DECERN introduces a multifaceted informativeness measure that combines discrepancy-confusion uncertainty and calibration diversity.
    The discrepancy-confusion uncertainty quantifies the structural stability and category directionality of fine-grained unlabeled data during local feature fusion.
    Subsequently, uncertainty-weighted clustering is performed to diversify the uncertainty samples.
    Then we calibrate the diversity to maximize the global diversity of the selected sample while maintaining its local representativeness.
    Extensive experiments conducted on 7 fine-grained image datasets across 39 distinct experimental settings demonstrate that our method achieves superior performance compared to state-of-the-art methods.
\end{abstract}


\section{Introduction}
The success of deep neural networks is highly dependent on large-scale annotated data.
However, a large amount of data is unlabeled, and obtaining high-quality data annotation is a time-consuming task that requires expert knowledge~\cite{LAL, ISO, EAOA}.
For fine-grained images, as well as images in specialized fields, such as archaeology, medicine, and natural species, the budget for expert annotation is even more expensive.
Active learning (AL)~\cite{atlas1989training, settles2009active, aggarwal2014active, ren2021survey, liu2022survey, zhan2022comparative, li2024survey, zha2025data, wan2024survey} methods have been proposed to achieve the construction of high-quality labeled datasets with limited budgets and maximize model training performance by iteratively selecting informative samples, rather than the entire data, to be annotated during the training process. 

Various AL methods have been proposed and can be broadly categorized into uncertainty-based, diversity-based, and hybrid methods. 
Uncertainty-based methods~\cite{ALFA,BALQUE,NoiseStability} select uncertain samples near decision boundaries for annotation as those that have the most information. 
Diversity-based methods~\cite{CoreSet_arxiv,CoreGCN,ActiveFT} select samples with maximum diversity to cover the distribution of unlabeled data without redundancy.
Hybrid methods~\cite{BADGE,CLUE,ALFA,CDAL} exploit both uncertainty and diversity.
These methods typically rely on reliable feature representations to estimate sample informativeness by analyzing the underlying data distribution~\cite{DOKT}.
However, in fine-grained image classification, where images represent subcategories under the same base category, visual similarity at shallow levels results in highly shared semantic features across deep network representations~\cite{ma2025optimal,wu2023bi,sun2018multi}. 
The overlap in feature distributions leads to unreliable uncertainty values estimates and reduces the effectiveness of diversity sampling, since samples from different subcategories may appear close in the feature space while being semantically distinct.

Recently, data augmentation methods~\cite{ALFA,ALMuLa-mix,DOKT} have been able to efficiently explore the neighborhood of unlabeled samples by constructing convex combinations of features through feature fusion, amplifying subtle differences between representations. 
This has led to approaches such as ALFA-Mix~\cite{ALFA}, which identify informative samples by evaluating the label variability of perturbed versions of unlabeled samples. 
However, the difficulty of changing labels varies across samples, and ALFA-Mix's approach of relying only on label variability to identify informative samples may overlook fine-grained samples that have the same pattern across multiple categories.

\begin{figure*}[t]
    \centering
    \includegraphics[
        width=0.99\textwidth,
        height=0.435\textheight,
    ]{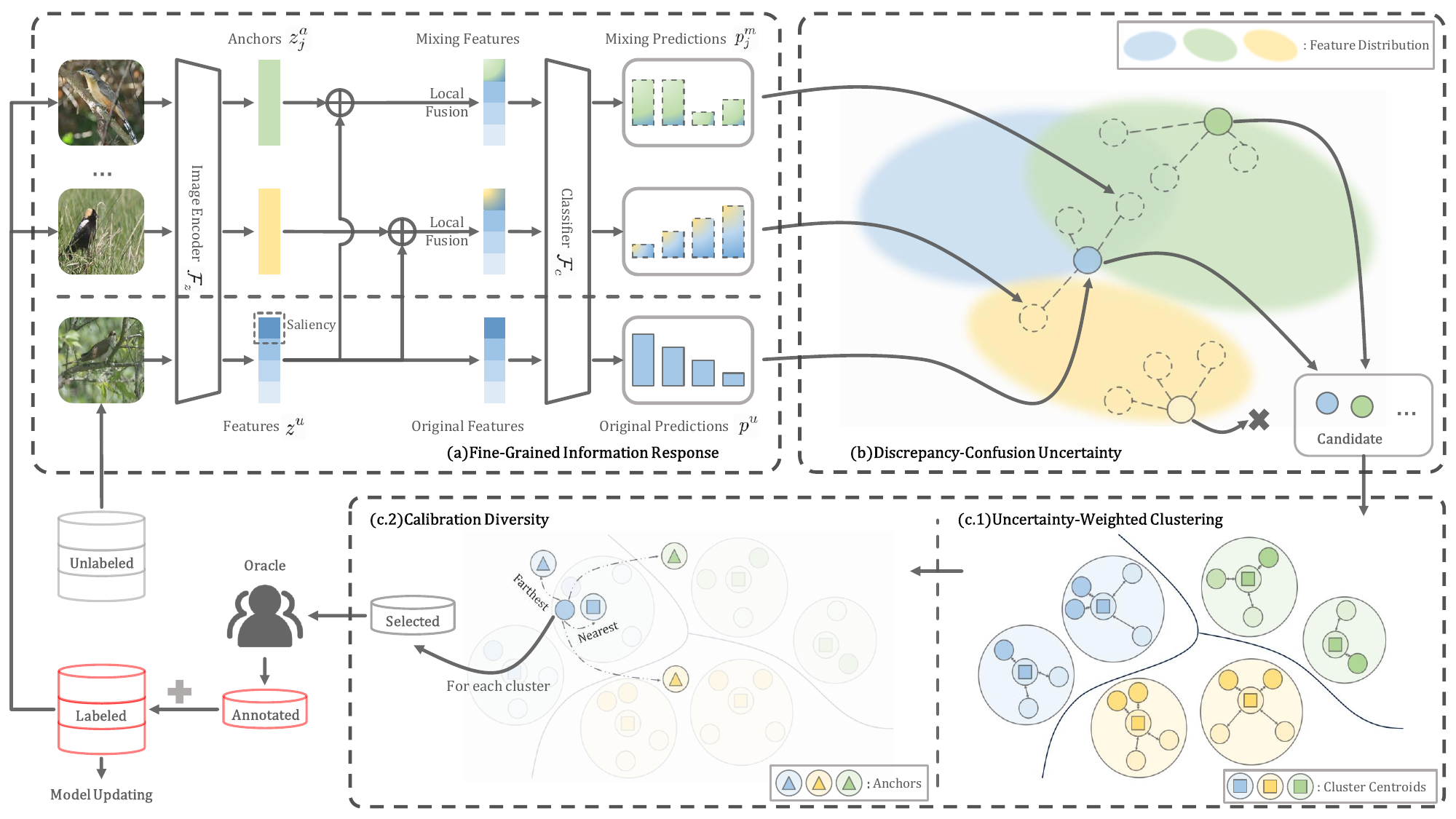}
    \caption{
        The overview of our DECERN framework.
        Our method \hyperref[sec:fusion]{(a)} comprehensively evaluates the variance of the probability distribution after the local feature fusion operations between the unlabeled data and the labeled data anchors.
        \hyperref[sec:score]{(b)} Unlabeled data with poor structural stability or low category directionality features
        are identified as high uncertainty samples and selected to form the candidates.
        After \hyperref[sec:clustering]{(c.1)} the uncertainty-weighted clustering, we further \hyperref[sec:clustering]{(c.2)} refine the selected uncertainty data by utilizing the calibration diversity, which achieves the trade-off between the local representativeness and the global diversity.
    }
    \label{fig:framework}
\end{figure*}

In this paper, we propose DECERN, a novel AL method for active fine-grained image classification.
As shown in \cref{fig:framework}, we first strategically deploy local feature fusion mechanisms to disentangle 
the structural stability and the category directionality of unlabeled samples.
Notably, low semantic-inconsistent perturbations trigger significant distributional shifts, exhibiting poor structural stability of features.
Meanwhile, high semantic-consistent perturbations induce knowledge confusion in samples, eroding the category directionality of features. 
Then the discrepancy-confusion uncertainty takes into account both the poor structural stability and the low category directionality.
Subsequently, we select candidates with high uncertainty scores from the unlabeled data.
After performing uncertainty-weighted clustering on the candidates, we utilize calibration diversity to enrich the diversity of the selected uncertainty samples.
On the one hand, we assume that samples proximal to uncertainty-weighted cluster centroids have larger local representativeness.
These instances containing prototypical features or critical patterns of unlabeled data serve to consolidate the model's comprehension of pivotal regions within the real distribution.
On the other hand, samples far from the labeled data centroids, called anchors, demonstrate more global diversity, exhibiting the greatest divergence from established knowledge, and enabling more effective exploration of regions such as decision boundaries and potential novel categories.

Our contributions are summarized as follows:
\begin{itemize}
    \item  
    We propose a novel active learning method, DECERN, for fine-grained image classification task.
    Our method efficiently constructs high-quality labeled data of the fine-grained image under limited annotation budgets, leading to significant performance gains.
    \item 
    We introduce a multifaceted informativeness measure in DECERN, combining discrepancy-confusion uncertainty and calibration diversity, to facilitate the effective selection of high informative samples.
    The former identifies features that exhibit poor structural stability and low category directionality during local feature fusion.
    The latter strategically balances the sample selection process, ensuring adequate local representativeness while maximizing global diversity.
    \item 
    We implement our proposal active learning strategy on 7 common-used fine-grained image classification datasets across 39 distinct experimental settings. 
    Extensive experimental results show that our DECERN achieves leading performance and outperforms existing state-of-the-art AL methods in the fine-grained image classification task.
\end{itemize}

\section{Related Work}
\textbf{Diversity-based methods}~\cite{ActiveFT,VAAL,TAVAAL,SRAAL,TypiClust,ProbCover} select a subset of unlabeled data that is representative of the entire pool. 
Specifically, the representative samples should be both broadly distributed and non-redundant.
A common strategy to achieve this diversity selection is to solve the coverage problem~\cite{MaxHerding,UHerding,CoreSet_arxiv,CoreGCN}.
For example, CoreSet~\cite{CoreSet_arxiv} frames active learning as a coreset selection problem, solved using the farthest-first traversal. 
Similarly, CoreGCN~\cite{CoreGCN} leverages graph node embeddings to differentiate labeled and unlabeled nodes, applying these representations within the CoreSet~\cite{CoreSet_arxiv} framework.
Other approaches utilize auxiliary models. 
ActiveFT~\cite{ActiveFT} directly optimizes a parametric model to select samples distributed similarly to the entire unlabeled data. 
Despite their effectiveness, diversity-based methods are limited~\cite{ren2021survey} by the abundance of categories within the pool and the constraints on the batch size.

\textbf{Uncertainty-based methods}~\cite{MIDL,LLAL,BUAL,VaGeRy,BalEntAcq,DBAL,DDU} select samples near the decision boundaries, which are most confusing to the task model.
Various indicators are used to estimate the uncertainty values of unlabeled data,
such as margin~\cite{Margin, balcan2007margin},
confidence~\cite{BALQUE},
entropy~\cite{EOAL,Entorpy2,Entorpy},
energy~\cite{EADA},
and influence~\cite{ISAL}.
In addition, some methods~\cite{LLAL,TiDAL,COVID-AL} use auxiliary modules to directly estimate the uncertainty values.
One of the most representative works, LLAL~\cite{LLAL} designs a loss prediction module to predict the target losses of unlabeled inputs. 
However, in fine-grained scenarios with high semantic similarity, overlapping feature distributions, and unreliable representations, uncertainty-based methods struggle to identify reliable decision boundaries~\cite{DOKT,BALQUE} and thus fail to guide the optimal selection of informative uncertainty samples.

\textbf{Hybrid methods}~\cite{BADGE,CLUE,ALFA,CDAL} combine the advantages of uncertainty and diversity.
For example, BADGE~\cite{BADGE} performs batch active learning by clustering gradient embeddings of the final output layer, which implicitly encode predictive uncertainty, to select diverse and informative unlabeled samples.
CLUE~\cite{CLUE} proposes cluster uncertainty-weighted embeddings to select diverse and informative target instances.
Although the hybrid criterion may reduce perceived informativeness, it significantly improves sample diversity through decreased inter-example similarity and achieves well performance.

\textbf{Feature fusion on active learning} has recently been considered in several studies~\cite{ALFA,ALMuLa-mix,DGMR,zhang2024employing}.
For example, ALFA-Mix~\cite{ALFA} constructs interpolations between labeled and unlabeled data representations and measures sample values based on the model's predicted inconsistency of pseudo-labels.
ALMULA-mix~\cite{ALMuLa-mix} improves on the original ALFA-Mix~\cite{ALFA} method, which is interpolated through weighted anchors to achieve a time-efficient identification of novel features. 
These methods well attempt the combination of feature fusion and active learning, exploring the neighborhood of unlabeled samples for efficient selection.

However, existing AL works predominantly focus on general image recognition or other scenarios, paying less attention to fine-grained images in specialized domains. 
Moreover, the integration of active learning with fine-grained image classification to achieve efficient annotation under limited budgets remains unexplored and constitutes the primary focus of this work.

\section{Method}
\subsection{Overview}
With a limited annotation budget, active learning aims to iteratively select the most informative samples from the unlabeled data pool for efficient annotation. 
Specifically, we have a large pool of unlabeled data \(\mathcal{D}^u=\{(x_i^u)\}_{i=1}^{N_u}\) and a pool of labeled data \(\mathcal{D}^\ell=\{(x_i^\ell,y_i^\ell)\}_{i=1}^{N_\ell}\) formed by random selection. 
We implement an active learning strategy that iteratively selects a subset of \(B\) samples from \(\mathcal{D}^u\) to be labeled by the oracle. 
Then we update \(\mathcal{D}^u\) and \(\mathcal{D}^\ell\) with the selected annotated samples and use the updated labeled data pool to train the neural network models \(\mathcal{F}=\mathcal{F}_z\circ \mathcal{F}_c\), where \(\mathcal{F}_z\) and \(\mathcal{F}_c\) denote the feature encoder and the classifier. 

\cref{fig:framework} illustrate our DECERN framework, and the pseudo-code is shown in Algorithm~\ref{code}.
We simultaneously consider two aspects: the discrepancy-confusion uncertainty and the calibration diversity of the samples, thus proposing two sampling strategies.
Specifically, for the discrepancy-confusion uncertainty sampling strategy,
we propose in \cref{sec:fusion} to achieve revealing the structural stability and category directionality of fine-grained features through local feature fusion.
Afterward, we define the discrepancy uncertainty and confusion uncertainty to evaluate these two types of features in \cref{sec:score}, and select samples with higher uncertainty to form candidates.
For the calibration diversity sampling strategy, we performed the uncertainty-weighted clustering operation to enrich the diversity of the selected samples in \cref{sec:clustering}.
In addition to selecting samples that are closest to the cluster centroids, we also require that their distance from the labeled data centroids, called anchors, is the farthest, which balances the local representativeness and global diversity.

\begin{algorithm}[t]
    \caption{Pseudo-code of DECERN}
    \label{code}
    \KwIn{Unlabeled data pool \(\mathcal{D}^u\), labeled data pool \(\mathcal{D}^\ell\), annotation budget \(B\), number of categories \(N_c\), feature encoder \(\mathcal{F}_z\), classifier \(\mathcal{F}_c\), local feature fusion strategy \(\phi\)}
    Construct labeled data anchors \(z^a\) and \(p^a\) based on the feature representation and prediction probability of \(\mathcal{D}^\ell\) by using \(\mathcal{F}_z\) and \(\mathcal{F}_c\)\;
    \For{\(x^u \in \mathcal{D}^u\)}{
        \(z^u=\mathcal{F}_z(x^u),\ p^u=\mathcal{F}_c(z^u)\)\;
        \For{\(j=1,2,...,N_c\)}{
            Perform local feature fusion \(\phi\) via
            \cref{eq:local_mix},
            and calculate the prediction probability of mixing representation \(p^m\) via
            \cref{eq:pm}\;
            Calculate the prediction probability \(p^b\) via
            \cref{eq:pb},
            and \(p^w\) via
            \cref{eq:pw}\;
    
            Calculate the category-level discrepancy-confusion uncertainty via
            \cref{eq:score_i},
            by using \(p^u\), \(p^b\), \(p^w\) and \(p^m\)\;
        }
        Calculate the instance-level discrepancy-confusion uncertainty score \(\mathcal{S}\) for unlabeled data via
        \cref{eq:score}\;
    }
    Select samples with high uncertainty scores \(\mathcal{S}\) as candidates \(\mathcal{C}\) by a dynamic threshold \(\zeta\) via
    \cref{eq:threshold}\;
    Perform uncertainty-weighted clustering on feature representations of the candidates \(\mathcal{C}\) and obtain \(B\) clusters \(\mathcal{C}_k\)\;
    Obtain the final selected batch as $\mathcal{D}^s=\{(x^{s_k})\}_{k=1}^B$, where one instance $x^{s_k}$ is selected from each cluster $\mathcal{C}_k$, and is closest to the cluster centroid \(z^{\mathcal{C}_k}\) and farthest to \(z^a\) via \cref{eq:diversity_sampling}\;
    \(y^{s_k}=Oracle(x^{s_k})\)\;
    $\mathcal{D}^{u} = \mathcal{D}^{u} \setminus \{(x^{s_k})\}_{k=1}^B$, $\mathcal{D}^{\ell} = \mathcal{D}^{\ell} \cup \{(x^{s_k}, y^{s_k})\}_{k=1}^B$\;
    Update the target model \(\mathcal{F}_z\) and \(\mathcal{F}_c\), and start the next AL cycle\;
\end{algorithm}

\subsection{Fine-Grained Information Response}
\label{sec:fusion}
With the feature encoder \(\mathcal{F}_z\) and the classifier \(\mathcal{F}_c\), we obtain the feature representations \(z^u\) and the prediction probabilities \(p^u\) of the unlabeled data:
\begin{equation}
    z^u=\mathcal{F}_z(x^u),\quad p^u=\mathcal{F}_c(z^u).
    \label{eq:zu-pu}
\end{equation}

We also compute the class anchors \(z^a\) and \(p^a\) by averaging the feature representation and the prediction probability of the labeled data for the \textit{j}-th category:
\begin{equation}
    \begin{split}
         &z^a_j=\frac{\sum_{(x,y)\in \mathcal{D}^\ell}\mathbb{1}\{y=j\}\cdot \mathcal{F}_z(x)}{\sum_{(x,y)\in \mathcal{D}^\ell}\mathbb{1}\{y=j\}}, \\
         &p^a_j=\frac{\sum_{(x,y)\in \mathcal{D}^\ell}\mathbb{1}\{y=j\}\cdot \mathcal{F}_c(\mathcal{F}_z(x))}{\sum_{(x,y)\in \mathcal{D}^\ell}\mathbb{1}\{y=j\}},
    \end{split}
    \label{eq:za-pa}
\end{equation}
where \(\mathbb{1}\{\cdot\}\) is an indicator function.

Then, we define the binary mask \(M\) via a pooled‑gradient scheme.
We first adaptively pool \(\mathcal{G}\) into 100 bins, then select the top‑\(\eta\) bins, and upsample the bin selection back to the original dimension.
The mask is thus given by
\begin{equation}
    M_{i,d} = 
    \begin{cases} 
    1 & \text{if dimension } d \text{ of sample } i\text{ falls into one of the top-}\eta\text{ bins},\\
    0 & \text{otherwise}.
    \end{cases}
    \label{eq:mask}
\end{equation}
where \(\mathcal{G} = \nabla_{z^u} \mathcal{L}(\mathcal{F}_c(z^u), \hat{y})\) denotes the backpropagation gradient of the unlabeled features, and \(\hat{y}\) is the pseudo‑label.
This pooled‑gradient mask suppresses noisy isolated dimensions and makes local fusion comparable across backbones.
As a result, \(M\) reduces the impact of irrelevant feature representations.

Once the binary mask \(M\) is constructed, the strength of the feature fusion \(\alpha\) requires specification.
Specifically, \(z^u\) tends to be similar to \(z^a\) in the confidence prediction category and not similar in other categories, due to their reliable and discriminative representation.
Consequently, leveraging insights from previous work~\cite{faramarzi2022patchup,ALFA},
we give large \(\alpha\) for each unlabeled data in their confidence prediction category and observe whether the mixing representation consistently maintains the category directionality under high semantic-consistent perturbations.
We propose that the novel features supporting the category are corrupted during the feature fusion process, which are conducive to calibrating the category boundary.
In contrast, for other categories, we give a small \(\alpha\), implement low semantic-inconsistent perturbations, to observe whether the fusion operation destroys the original semantic structure.
Learning about features of unstable structures promotes robust feature representations and contributes to the consistent recognition of data from the same fine-grained category.
Therefore, we set \(\alpha_j\) to \(p^u_j\).

Then, we perform the local feature fusion strategy \(\phi\) on unlabeled data \(z^u\) and previously obtained category anchors \(z^a_j\) for the \textit{j}-th category:
\begin{equation}
    \begin{split}
        \phi(z^u,z^a_j;\alpha_j,M)=(1 - M) z^u + M \left(z^u(1 - \alpha_j) + z^a_j\alpha_j\right).
    \end{split}
    \label{eq:local_mix}
\end{equation}

Finally, we assess the response of unlabeled samples under varying feature fusion operations using the following indicators:
Prediction probability of the original representation \(p^u\); 
Fusion prediction probability \(p^b\) that measures the theoretical prediction probability for feature fusion data;
Weighted prediction probability \(p^w\) that measures the theoretical prediction probability for local feature fusion data;
Prediction probability of mixing representations \(p^m\) that measures probability offsets for local feature fusion data;
Specifically, \(p^u\) is defined in \cref{eq:zu-pu} and others are formulated as:
\begin{align}
    p^b_j&=(1 - \alpha_j) p^u + \alpha_j p^a_j,
    \label{eq:pb} \\
    p^w_j&=(1 - \eta \alpha_j) p^u + \eta \alpha_j p^a_j,
    \label{eq:pw} \\
    p^m_j&=\mathcal{F}_c(\phi(z^u,z^a_j;\alpha_j,M)),
    \label{eq:pm}
\end{align}
where \(\eta\) is the size of the mask \(M\).

These predicted probabilities capture the relationship between unlabeled data and category-specific decision boundaries, enabling the calculation of reliable uncertainty scores for each unlabeled sample.

\subsection{Discrepancy-Confusion Uncertainty}
\label{sec:score}
Our DECERN select samples that exhibit a higher discrepancy-confusion uncertainty.
When the local feature fusion operation corrupts the semantics of the original representation, resulting in large probability offsets, we should give priority to selecting these samples due to their high discrepancy uncertainty.
However, relying solely on discrepancy uncertainty prevents the selection of samples that are cross-category ambiguous.
To address this problem, we also exploit the confusion uncertainty of the fusion operation.
We utilize cross entropy \(\mathcal{CE}\) as discrepancy uncertainty \(\mathcal{S}_d\) to evaluate structural stability, and entropy \(\mathcal{H}\) as confusion uncertainty \(\mathcal{S}_c\) to evaluate category directionality.
After performing the local feature fusion operation of unlabeled data with various anchors,
the discrepancy-confusion uncertainty then identifies mixing representations of each operation by low discriminative power due to poor structural stability from ambiguous semantics or lack of differential signals (low category directionality). 
Unlabeled data whose representations exhibit these characteristics are prioritized.
Formally, we formulate the category-level discrepancy-confusion uncertainty score \(S_{dc}\) for the \textit{j}-th category as follows:
\begin{equation}
  \begin{split}
        \mathcal{S}_{dc}(p_j^*,p^m_j;\beta_j)
        =\mathcal{H}(p^m_j)^{1-\beta_j}+\mathcal{CE}(p^*_j,p^m_j)^{\beta_j},
  \end{split}
  \label{eq:score_i}
\end{equation}
where \(p_j^*\in{\{p^u,p^b_j,p^w_j\}}\), \(\beta_j=1-\frac{1+\cos(z^u,z^a_j)}{2}\), and \(\cos(\cdot,\cdot)\) is the cosine similarity.

Then, we calculate the instance-level discrepancy-confusion uncertainty score \(\mathcal{S}\) for each unlabeled data:
\begin{equation}
     \begin{split}
         \mathcal{S}=\frac{1}{N_c} \sum_{j=1}^{N_c} \left[(1-\eta) \mathcal{S}_{dc}(p^u,p^m_j;\beta_j)+
         \eta \mathcal{S}_{dc}(p^b_j,p^m_j;\beta_j))+ \mathcal{S}_{dc}(p^w_j,p^m_j;\beta_j)\right],
     \end{split}
     \label{eq:score}
\end{equation}
where \(N_c\) is the number of categories and \(\eta\), denoted in \cref{sec:fusion}, is the size of the mask \(M\).
This facilitates the precise identification of ambiguous samples situated near decision boundaries, thereby enhancing the model's ability to distinguish challenging samples.

Finally, we use reliable uncertainty scores \(\mathcal{S}\) and a dynamic threshold \(\zeta\) with the AL process for uncertainty-based sampling:
\begin{equation}
    \zeta =\bar{\mathcal{S}}+ \lambda \sigma \gamma,
    \ \sigma = \sqrt{ \mathbb{E}\big[ (\mathcal{S} - \bar{\mathcal{S}})^2 \big] },
    \ \gamma = \frac{\mathbb{E}\left[(\mathcal{S} - \bar{\mathcal{S}})^3\right]}{\sigma^3},
    \label{eq:threshold}
\end{equation}
where \(\bar{\mathcal{S}}\), \(\sigma\) and \(\gamma\) are the mean, standard deviation, and the skewness of \(\mathcal{S}\), \(\lambda\) is the moderator of uncertainty sampling intensity. 

We expect to select more high-uncertainty samples at the decision boundaries as model performance improves. 
Therefore, we set a dynamic threshold \(\zeta\), which is obtained by calculating the skewness of the scores and implies information about the improvement in model performance and the uncertainty scores of the samples.
And with the threshold \(\zeta\), we select data with scores \(\mathcal{S}\) above it as candidates \(\mathcal{C}\), which contain more uncertainty information.

\subsection{Uncertainty-Weighted Clustering and Calibration Diversity}
\label{sec:clustering}
The diversity of the samples is inappropriately neglected, especially when the decision boundaries are ambiguous and the uncertainty information is noisy.
Thus, we perform diversity clustering in feature representations to sample from diversity regions in the feature space, which is inspired by several AL methods~\cite{CEC,ALFA,CLUE}.
We assign weights to feature representation of each sample based on their uncertainty score \(\mathcal{S}\).
Intuitively, this particular weighting mechanism renders the cluster centroids more determinate by the position of the high-uncertainty samples, and enhances the potential of selection from them.
By implementing uncertainty-weighted clustering, we maintain the balance between uncertainty and diversity, avoiding the oblivion of pure diversity sampling to data uncertainty.

We use the K-Means algorithm for uncertainty-weighted clustering and obtain \(B\) cluster \(\mathcal{C}_k\).
Then we select the samples that are closest to the cluster centroids \(z^{\mathcal{C}_k}\) directly.
However, boundary samples and potential subcategory samples may be ignored even though cluster centroids are biased towards high-uncertainty regions.
To address this problem, we exploit the prior information from the anchors to calibrate the sample diversity.
For each cluster \(\mathcal{C}_k\), samples closest to \(z^{\mathcal{C}_k}\) and farthest away from \(z^a\) are selected as the subset with the AL strategy,
which can be formulated as:
\begin{equation}
    \begin{split}
        x^{s_k} = \arg\max_{x_i \in \mathcal{C}_k}\left[-\xi (1-\cos(z_i^u, z^{\mathcal{C}_k})) + (1-\xi) \min_{j}(1-\cos(z_i^u, z_j^a))\right],
    \end{split}
    \label{eq:diversity_sampling}
\end{equation}
where \(\xi\) is a hyperparameter, \(k=1,2,...,B\), and \(j=1,2,...,N_c\).

This function integrates two terms designed for synergistic sample selection: local representativeness and global diversity.
Crucially, local representativeness selects prototypical samples from high-uncertainty regions as typical representatives of intra-class core variations, thereby avoiding the selection of meaningless redundant samples within the class.
The global diversity identifies samples exhibiting maximal divergence from established knowledge to supplement the feature distributions.
This not only corrects for systematic gaps in the model's comprehension of the data distribution but also drives it to learn tail features and extend its decision boundaries.

Finally, we query the labels of these samples. 
After updating \(\mathcal{D}^\ell\), we train the task model on the labeled data.

\section{Experiments}
\subsection{Experiment Settings}
\textbf{Dataset and Metrics.}
We conduct experiments on 7 fine-grained image classification datasets:
Caltech101~\cite{exam_Caltech101},
BronzeDing~\cite{exam_BronzeDing},
CUB~\cite{exam_CUB},
Flowers102~\cite{exam_Flowers102},
Food101~\cite{exam_Food101},
OxfordIIITPet~\cite{exam_OxfordIIITPet}, and
StanfordDogs~\cite{exam_StanfordDogs}.
The raw training data serves as the unlabeled pool for selection.
We employ the \textit{Top-1 Accuracy} metric for performance evaluation.
Additionally, we also report the experimental results using the macro F1-score metric in Appendix D.3 to address the possible imbalance of the testing data.

\textbf{Baselines.} 
We compare DECERN with \textbf{Random},
\textbf{K-Means},
\textbf{Margin}~\cite{Margin},
\textbf{CoreSet}~\cite{CoreSet_arxiv},
\textbf{BADGE}~\cite{BADGE},
\textbf{NoiseStability}~\cite{NoiseStability},
\textbf{UHerding}~\cite{UHerding},
\textbf{CLUE}~\cite{CLUE},
\textbf{ALFA-Mix}~\cite{ALFA}. 
Additionally, we further compare it with \textbf{Confidence},
\textbf{Entropy}~\cite{Entorpy},
\textbf{CoreGCN}~\cite{CoreGCN},
\textbf{ActiveFT}~\cite{ActiveFT},
\textbf{BALQUE}~\cite{BALQUE},
\textbf{DropQuery}~\cite{DropQuery} in Appendix D.4 to validate the effectiveness of our method.
In summary, a total of 15 representative and recent AL baselines, covering uncertainty-, diversity-, hybrid-, and feature-fusion-based methods, including the closest ones, CLUE and ALFA-Mix, are compared with DECERN.

\begin{figure*}[t]
    \centering
    \begin{subfigure}[b]{0.24\textwidth}
        \centering
        \includegraphics[
            width=\textwidth, 
            height=0.139\textheight,
        ]{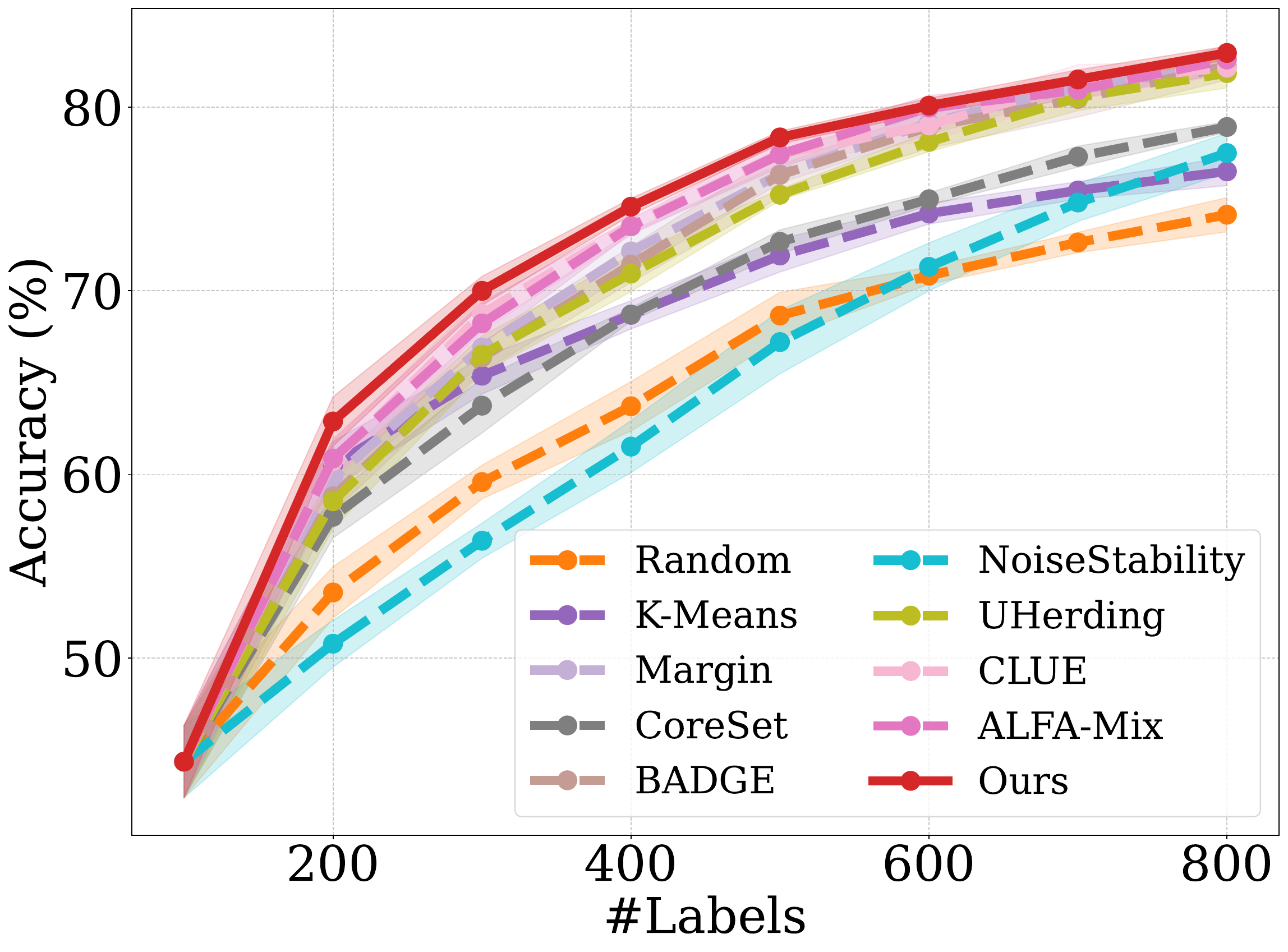}
        \caption{Caltech101, $K=1$}
        \label{fig:main_exp_R0107}
    \end{subfigure}
    \hfill
    \begin{subfigure}[b]{0.24\textwidth}
        \centering
        \includegraphics[
            width=\textwidth, 
            height=0.139\textheight,
        ]{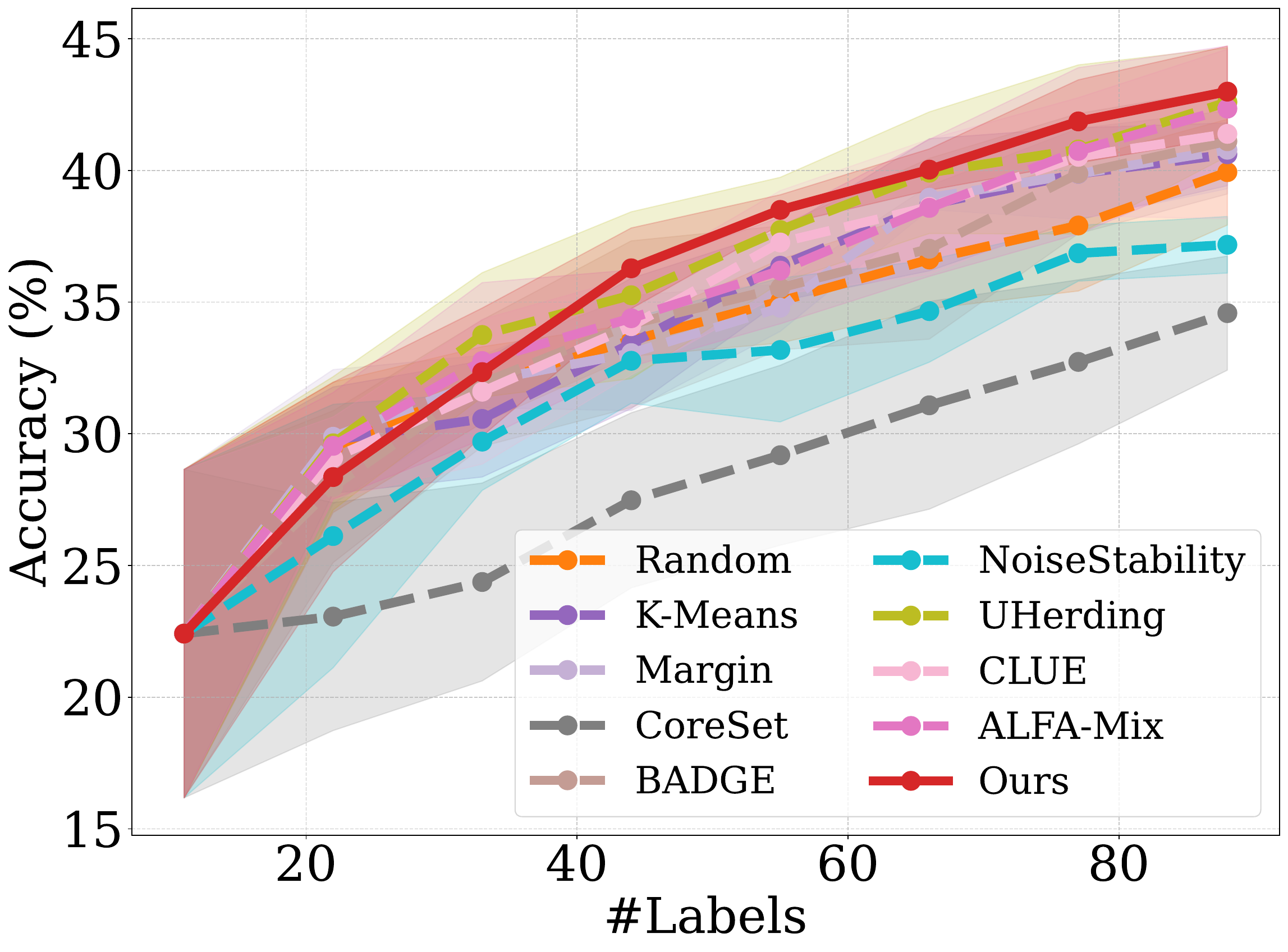}
        \caption{BronzeDing, $K=1$}
        \label{fig:main_exp_R0106}
    \end{subfigure}
    \hfill
    \begin{subfigure}[b]{0.24\textwidth}
        \centering
        \includegraphics[
            width=\textwidth, 
            height=0.139\textheight,
        ]{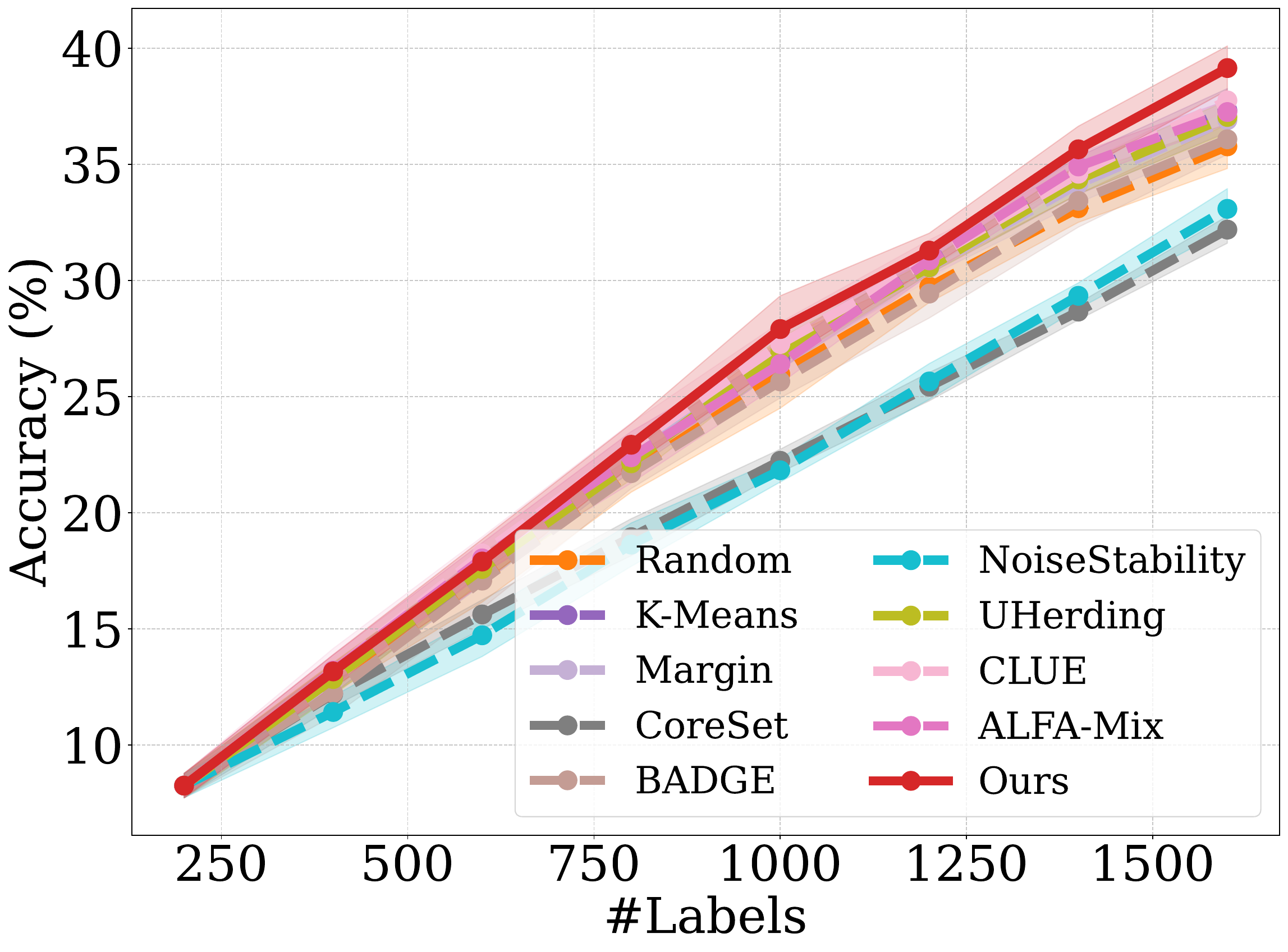}
        \caption{CUB, $K=1$}
        \label{fig:main_exp_R0101}
    \end{subfigure}
    \hfill
    \begin{subfigure}[b]{0.24\textwidth}
        \centering
        \includegraphics[
            width=\textwidth, 
            height=0.139\textheight,
        ]{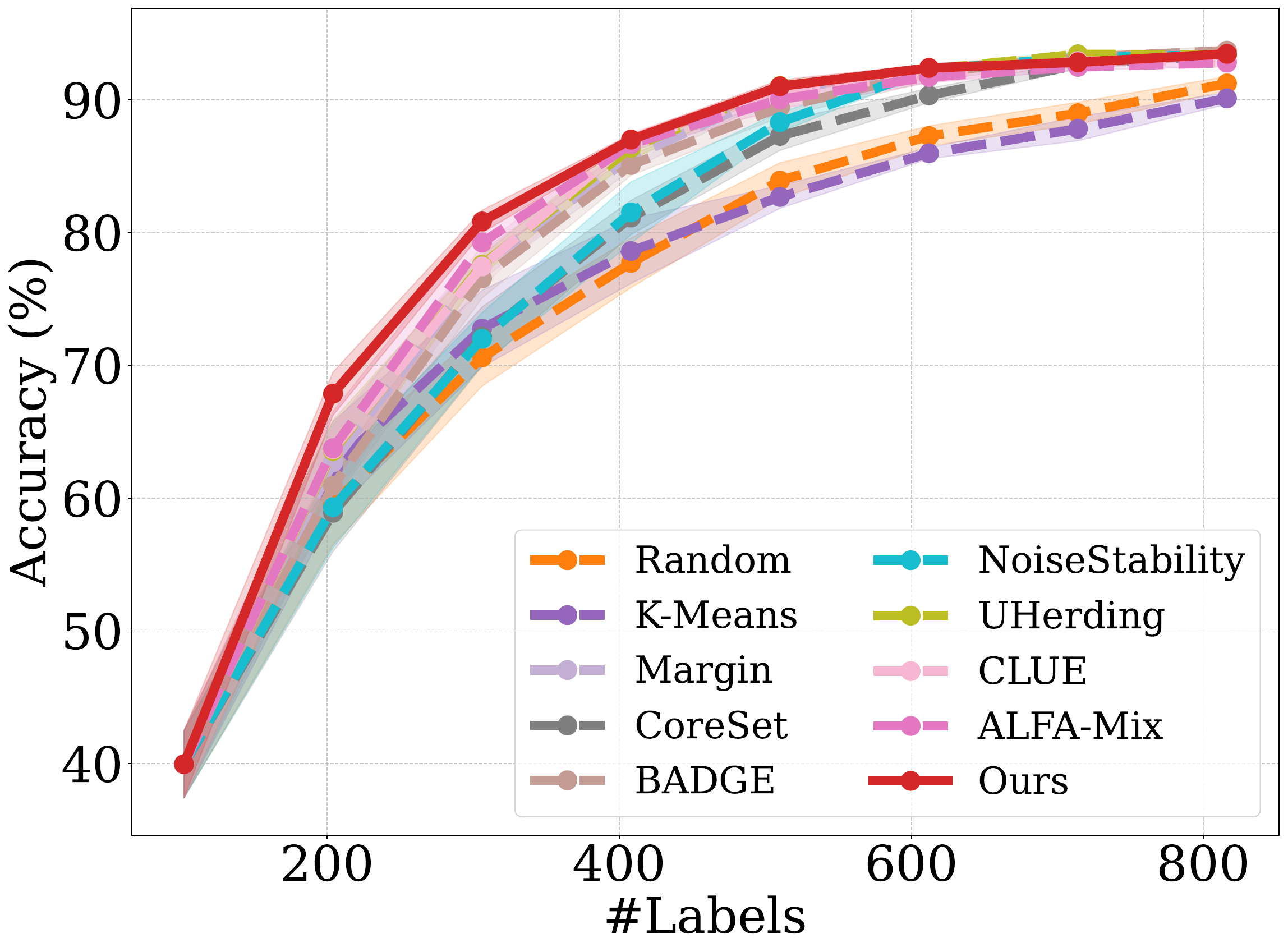}
        \caption{Flowers102, $K=1$}
        \label{fig:main_exp_R0109}
    \end{subfigure}

    \begin{subfigure}[b]{0.08\textwidth}
        \centering
        \includegraphics[
            width=\textwidth, 
            height=0.139\textheight,
        ]{static/WHITE.pdf}
    \end{subfigure}
    \begin{subfigure}[b]{0.26\textwidth}
        \centering
        \includegraphics[
            width=\textwidth, 
            height=0.139\textheight,
        ]{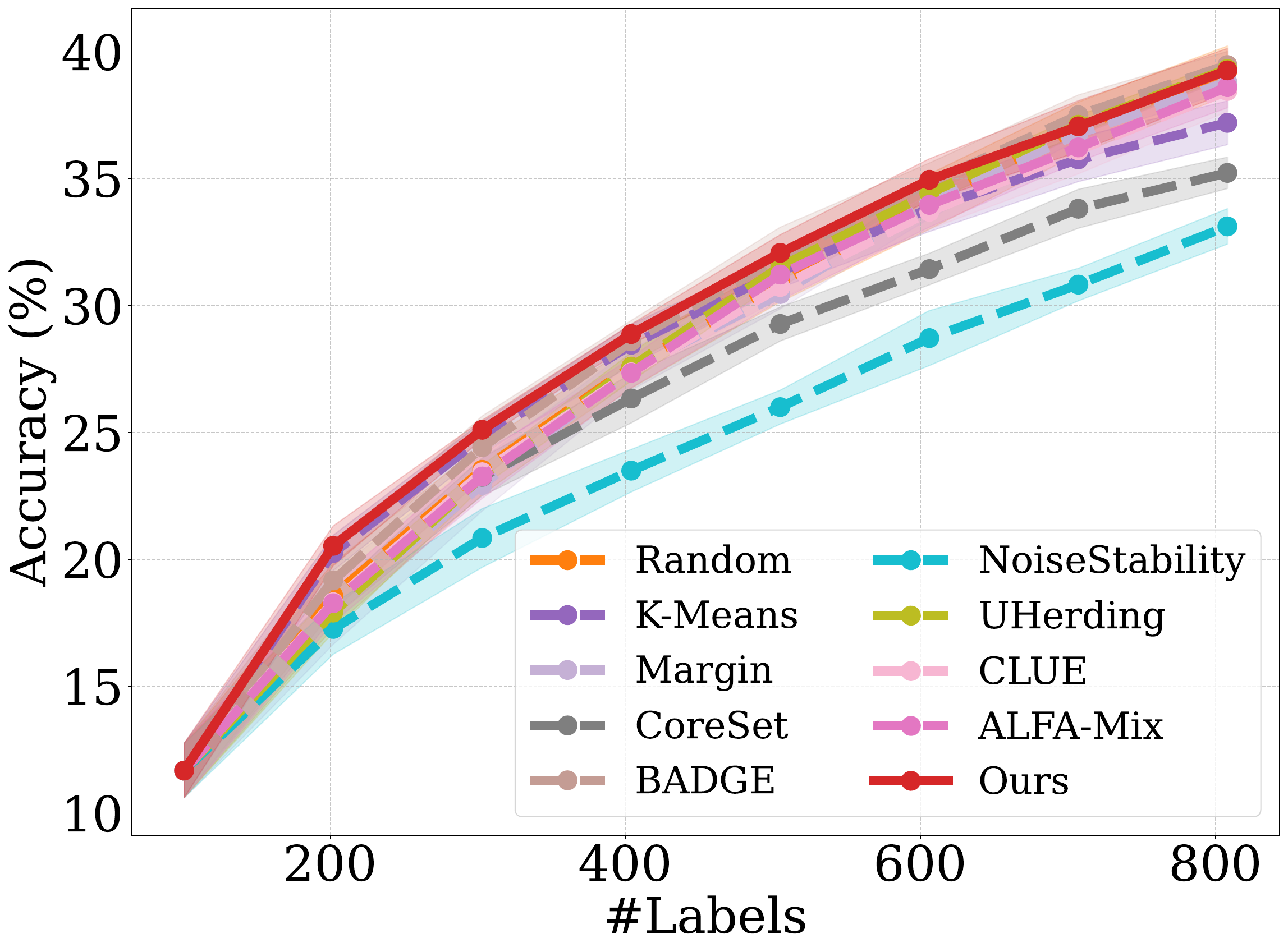}
        \caption{Food101, $K=1$}
        \label{fig:main_exp_10205}
    \end{subfigure}
    \hfill
    \begin{subfigure}[b]{0.26\textwidth}
        \centering
        \includegraphics[
            width=\textwidth, 
            height=0.139\textheight,
        ]{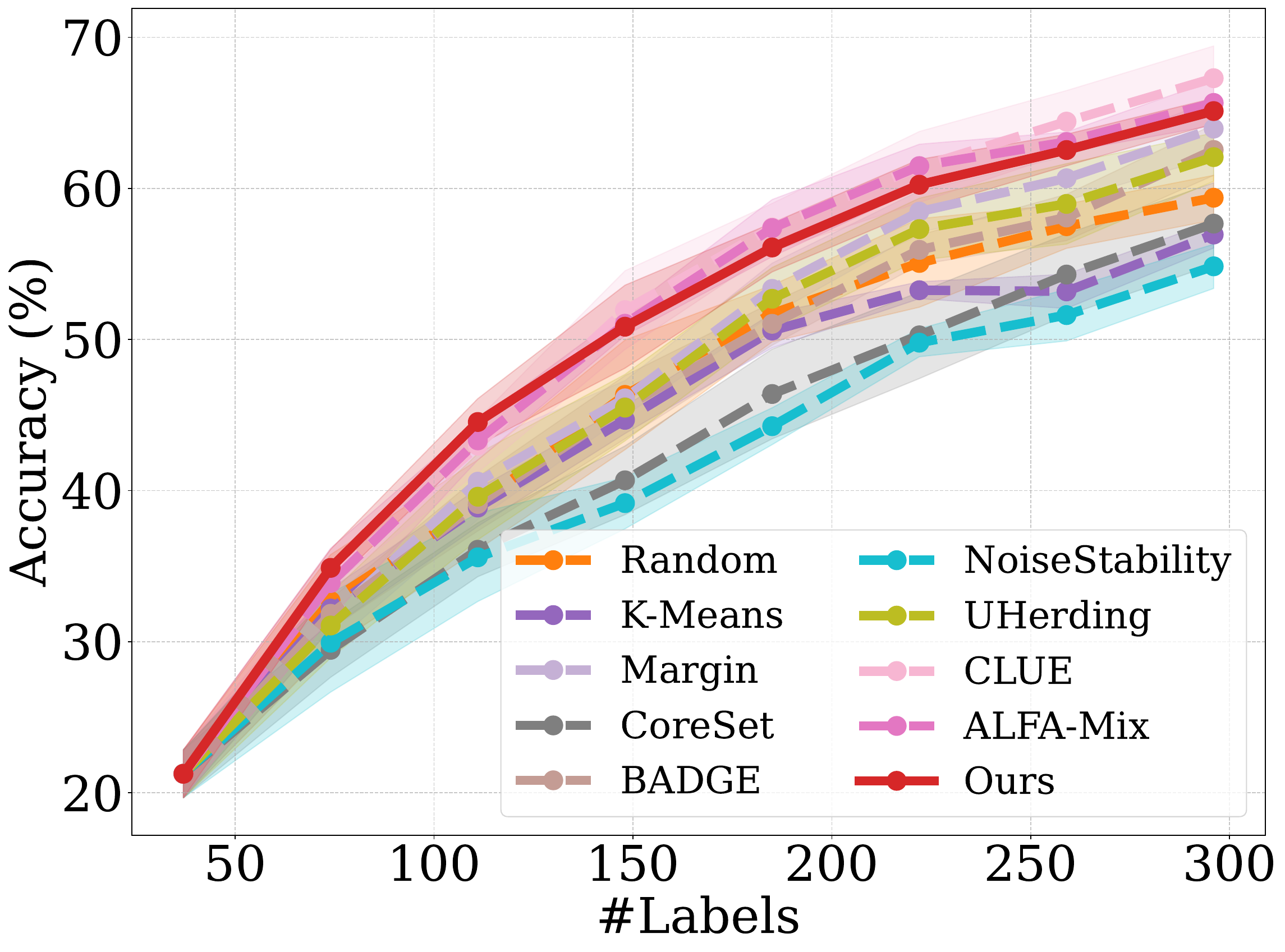}
        \caption{OxfordIIITPet, $K=1$}
        \label{fig:main_exp_R0100}
    \end{subfigure}
    \hfill
    \begin{subfigure}[b]{0.26\textwidth}
        \centering
        \includegraphics[
            width=\textwidth, 
            height=0.139\textheight,
        ]{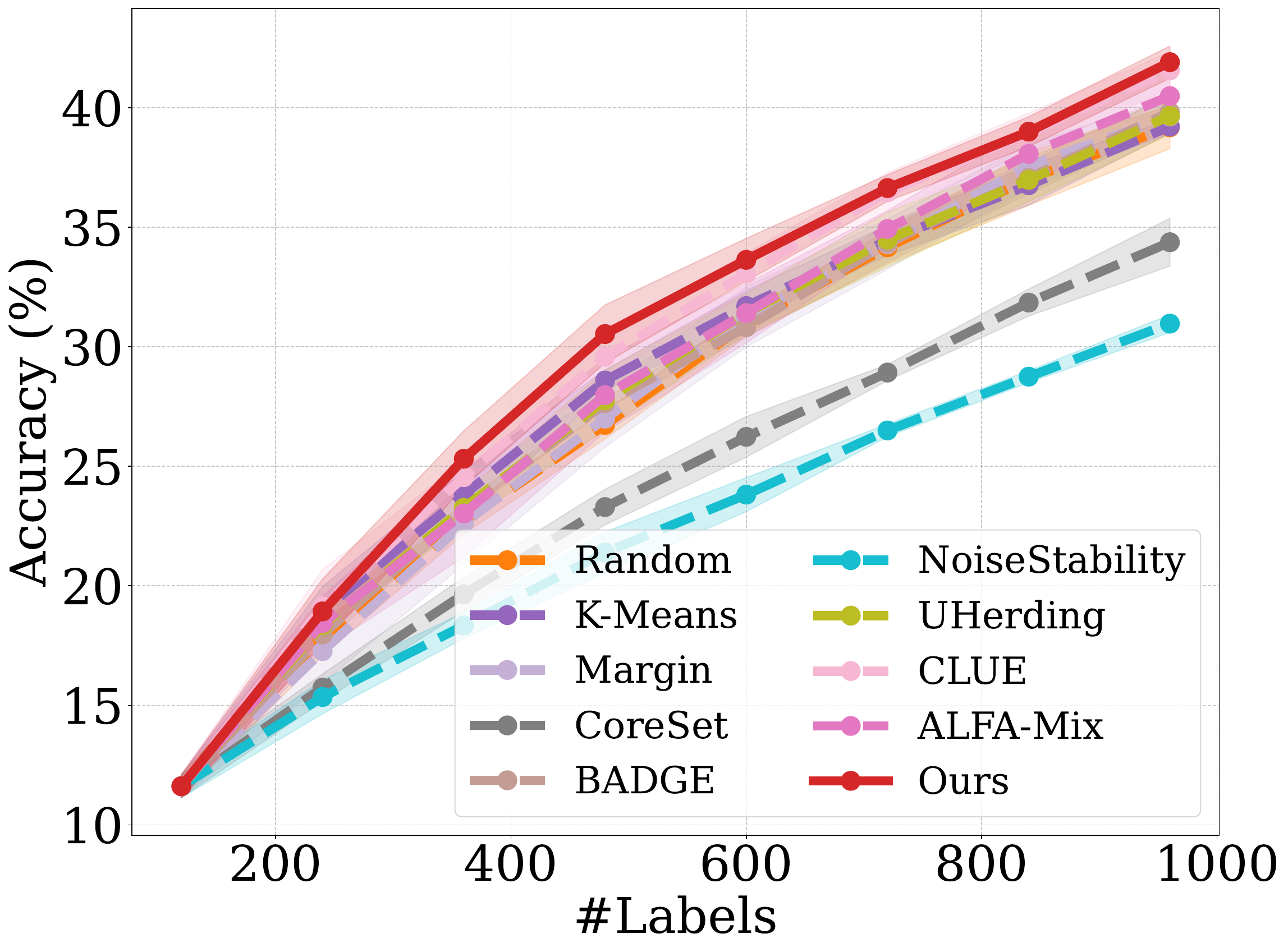}
        \caption{StanfordDogs, $K=1$}
        \label{fig:main_exp_R0108}
    \end{subfigure}
    \begin{subfigure}[b]{0.05\textwidth}
        \centering
        \includegraphics[
            width=\textwidth, 
            height=0.139\textheight,
        ]{static/WHITE.pdf}
    \end{subfigure}
    
    \begin{subfigure}[b]{0.08\textwidth}
        \centering
        \includegraphics[
            width=\textwidth, 
            height=0.139\textheight,
        ]{static/WHITE.pdf}
    \end{subfigure}
    \begin{subfigure}[b]{0.26\textwidth}
        \centering
        \includegraphics[
            width=\textwidth, 
            height=0.139\textheight,
        ]{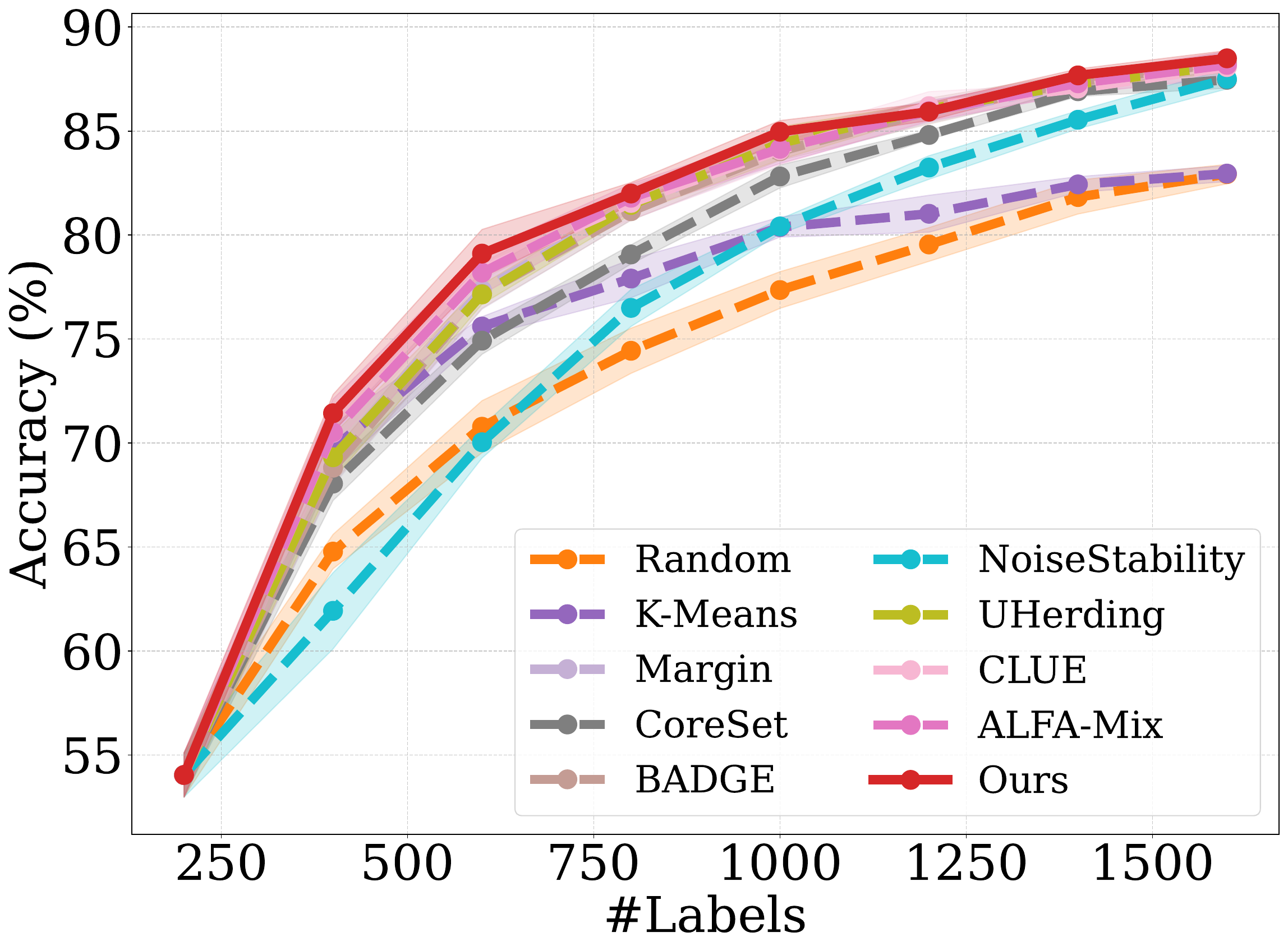}
        \caption{Caltech101, $K=2$}
        \label{fig:main_exp_R0207}
    \end{subfigure}
    \hfill
    \begin{subfigure}[b]{0.26\textwidth}
        \centering
        \includegraphics[
            width=\textwidth, 
            height=0.139\textheight,
        ]{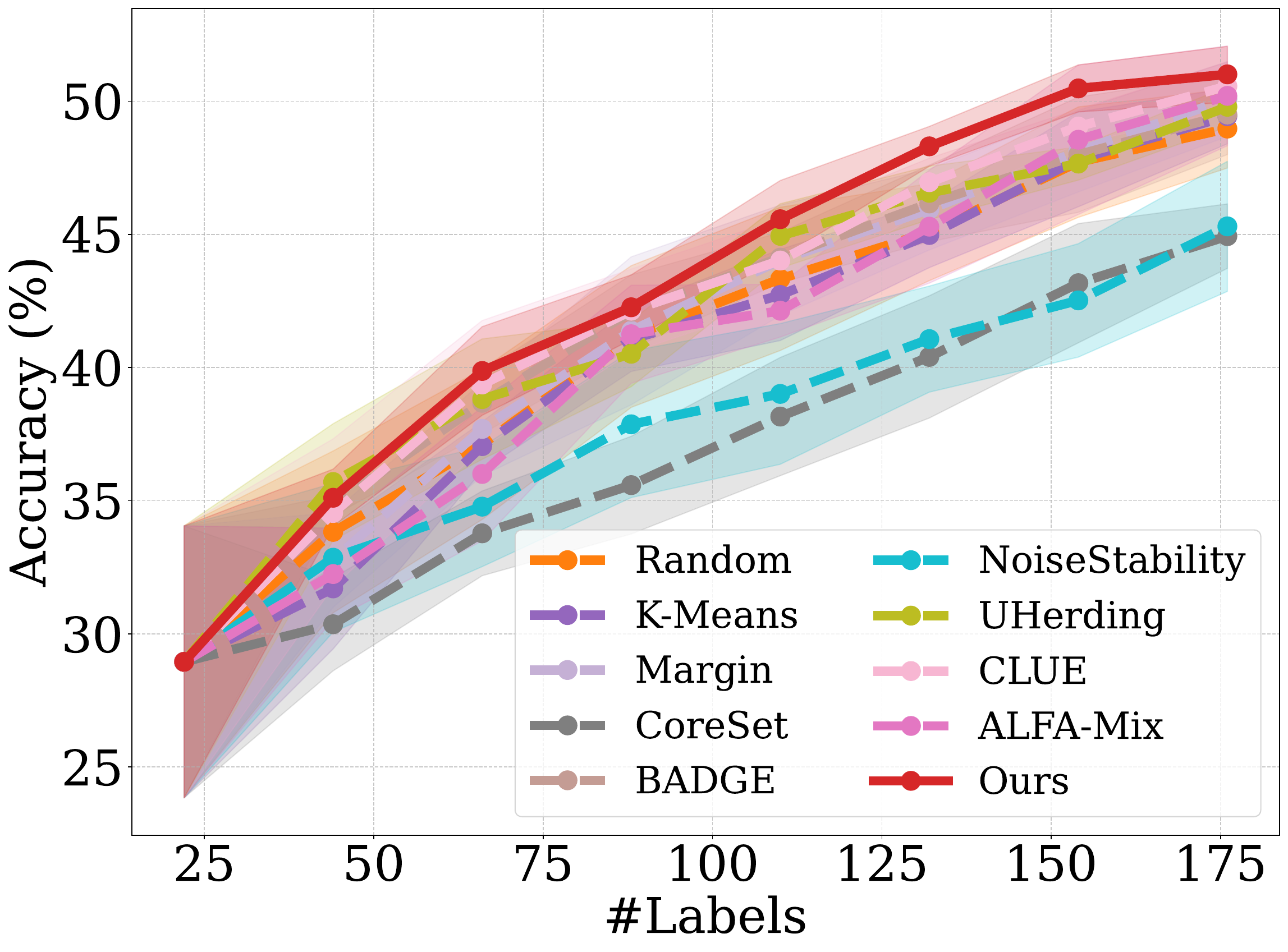}
        \caption{BronzeDing, $K=2$}
        \label{fig:main_exp_R0206}
    \end{subfigure}
    \hfill
    \begin{subfigure}[b]{0.25\textwidth}
        \centering
        \includegraphics[
            width=\textwidth, 
            height=0.139\textheight,
        ]{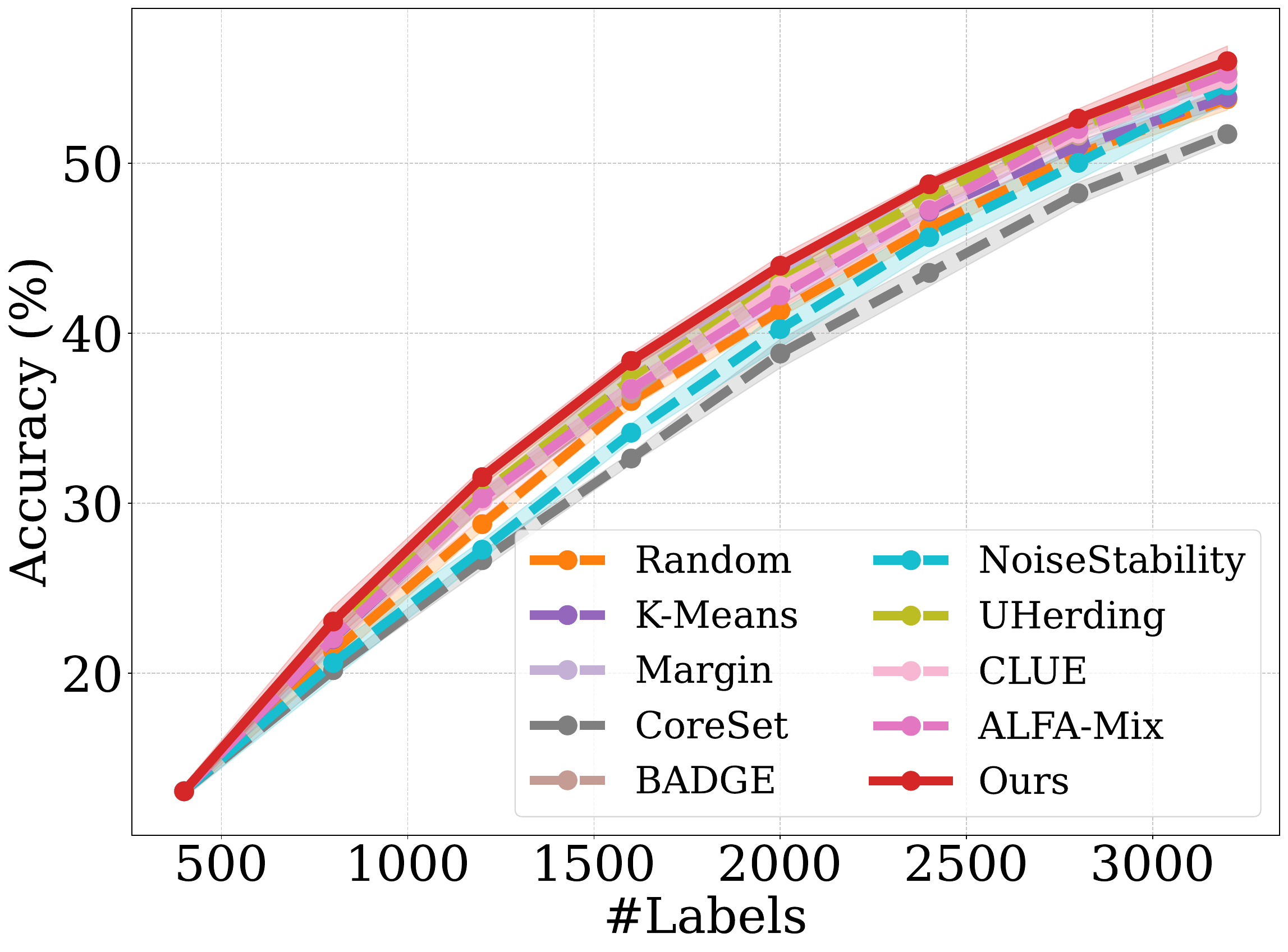}
        \caption{CUB, $K=2$}
        \label{fig:main_exp_R0201}
    \end{subfigure}
    \begin{subfigure}[b]{0.06\textwidth}
        \centering
        \includegraphics[
            width=\textwidth, 
            height=0.139\textheight,
        ]{static/WHITE.pdf}
    \end{subfigure}

    \begin{subfigure}[b]{0.08\textwidth}
        \centering
        \includegraphics[
            width=\textwidth, 
            height=0.139\textheight,
        ]{static/WHITE.pdf}
    \end{subfigure}
    \begin{subfigure}[b]{0.26\textwidth}
        \centering
        \includegraphics[
            width=\textwidth, 
            height=0.139\textheight,
        ]{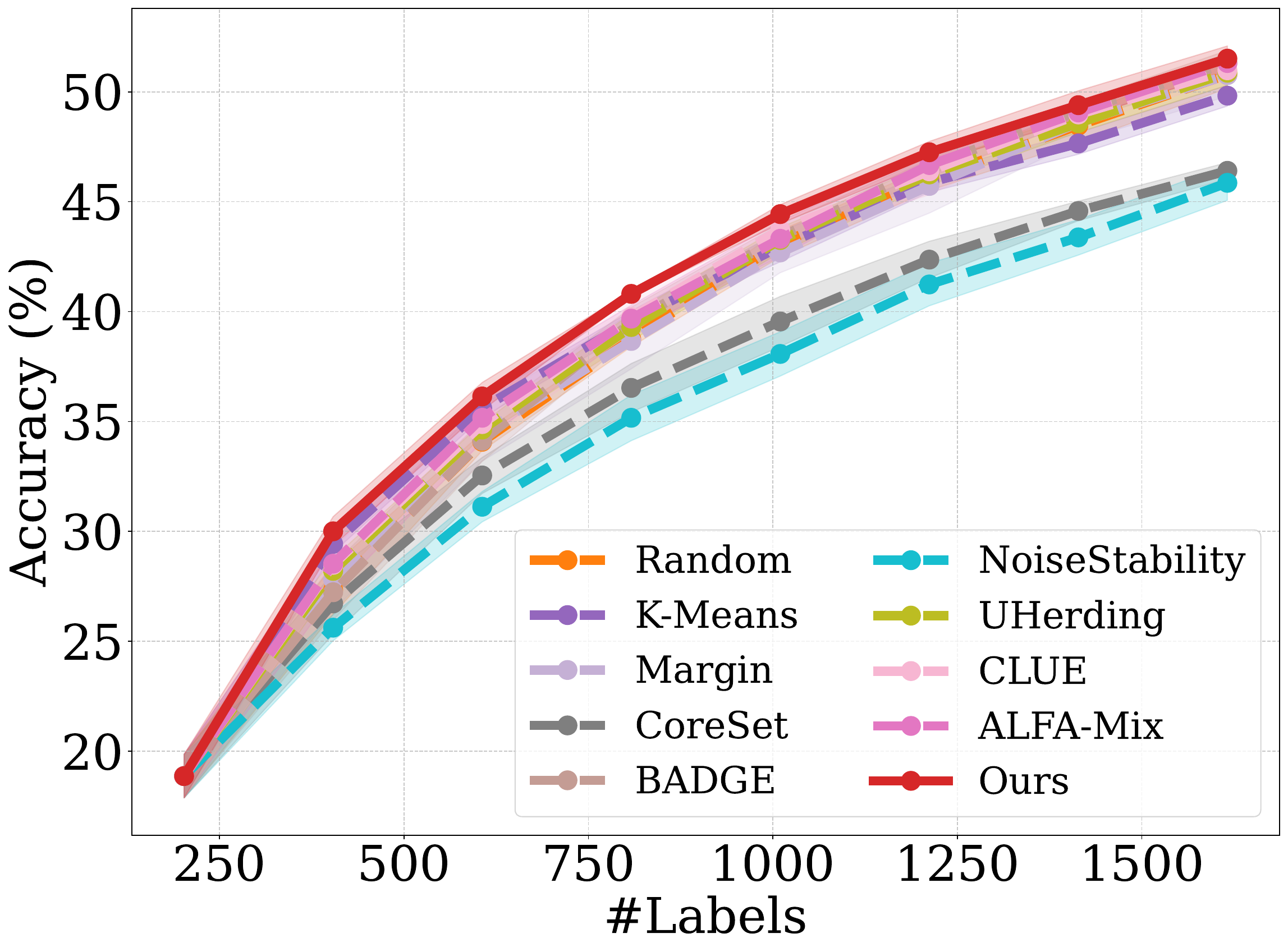}
        \caption{Food101, $K=2$}
        \label{fig:main_exp_R0205}
    \end{subfigure}
    \hfill
    \begin{subfigure}[b]{0.26\textwidth}
        \centering
        \includegraphics[
            width=\textwidth, 
            height=0.139\textheight,
        ]{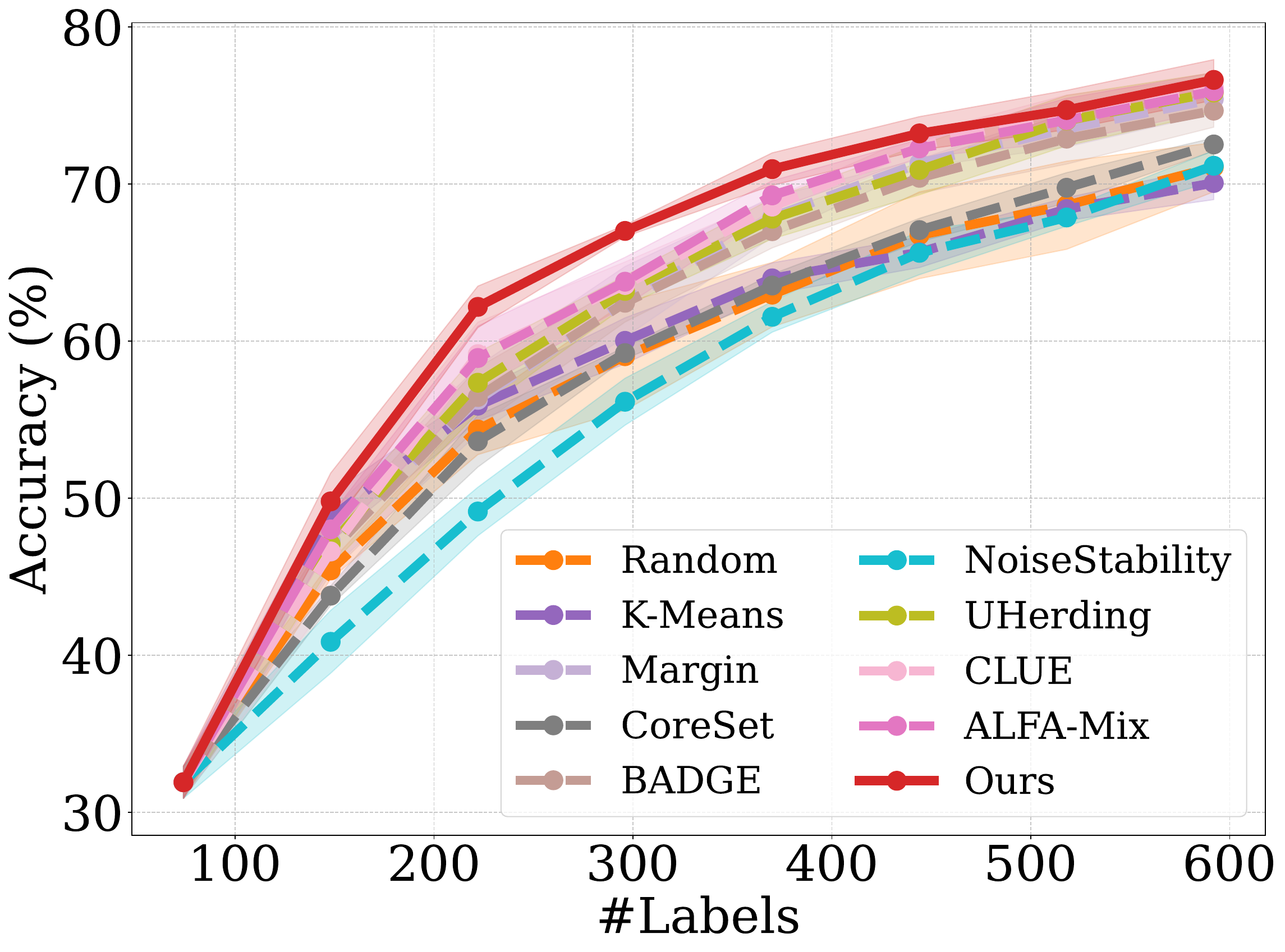}
        \caption{OxfordIIITPet, $K=2$}
        \label{fig:main_exp_R0200}
    \end{subfigure}
    \hfill
    \begin{subfigure}[b]{0.26\textwidth}
        \centering
        \includegraphics[
            width=\textwidth, 
            height=0.139\textheight,
        ]{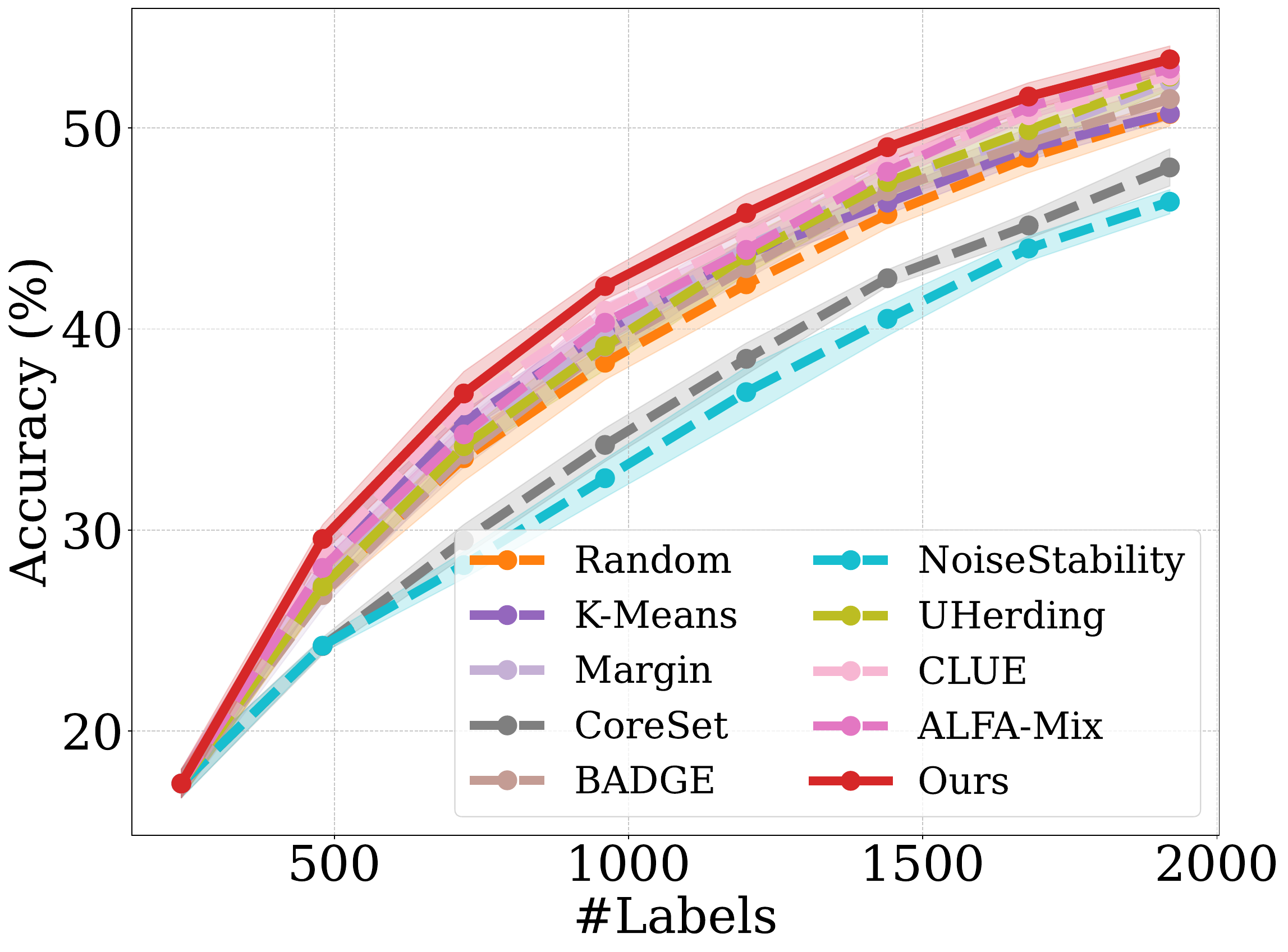}
        \caption{StanfordDogs, $K=2$}
        \label{fig:main_exp_R0208}
    \end{subfigure}
    \begin{subfigure}[b]{0.05\textwidth}
        \centering
        \includegraphics[
            width=\textwidth, 
            height=0.139\textheight,
        ]{static/WHITE.pdf}
    \end{subfigure}
    
    \caption{
        Comparison on benchmark datasets using ResNet50 model. 
        We use annotation budget $B=K\cdot N_c$ in each AL cycle.
        $N_c$ is the number of categories, $K\in \{1,2\}$.
    }
    \label{fig:main_exp_R02}
\end{figure*}

\textbf{Implementation Details.}
We adopt ResNet50~\cite{exam_resnet} or ViT-Small~\cite{exam_vit} pretrained by DINO~\cite{exam_dino} and DINOv2~\cite{oquab2023dinov2} as the backbone.
Throughout training, we employ the Adam~\cite{exam_adam} optimizer with a learning rate of 0.001 and cosine learning rate decay.
A consistent batch size of 128 is maintained in all experiments.
The number of AL cycles is set to 8.
In each AL cycle, we apply the sampling strategy to the entire pool of unlabeled data \(\mathcal{D}^u\),
except for the Food101 dataset, where we apply the AL strategy to a random subset.
In each AL cycle, we also select a subset of unlabeled data for labeling whose size is \(B=K\cdot N_c\) (\(N_c\) is the number of categories, \(K\in \{1,2\}\)),
except for the Flowers102 dataset, where we select a subset of size \(B=1\cdot N_c\).
Since the initial labeled data are required for the AL methods, we employ random sampling in the first AL cycle.

For robustness comparison, we repeat each experiment with 5 different seeds and report the mean and standard deviations of the results.
Meanwhile, we conduct extensive experiments involving multiple common fine-grained image datasets, different architectures, and varying annotation budgets of active learning (39 distinct experimental settings in total, and more details are provided in Appendix D.1.
All experiments are conducted on an NVIDIA A40 GPU with PyTorch~\cite{exam_torch}.
We generally set the values of \(\eta\), \(\lambda\), \(\xi\) at 0.1, 0.5, and 0.8, respectively, as these values perform well in all experiments.

\subsection{Overall Results}
\label{sec:res}
\textbf{Generality of our proposal DECERN.}
In \cref{fig:main_exp_R02}, we present the results over different AL cycles on benchmark datasets, using the ResNet50 model.
Appendix D.2 further provides results with the ViT-Small model pretrained using DINO and DINOv2 on the same benchmark datasets. 
Additionally, Appendix D.4 includes results with additional baselines to further demonstrate the effectiveness of our method.
In \cref{fig:main_exp_R02}, we can observe that our method outperforms other baseline methods in various settings, demonstrating the effectiveness and superiority of our approach.
Diversity methods (\eg, K-Means and CoreSet) and uncertainty methods (\eg, NoiseStability) are significantly hindered by fine-grained examples, as these examples lack discriminative features to distinguish between similar categories.
In contrast, hybrid strategies achieve competitive performance, but are still not optimal.

\textbf{Sampling imbalance affect the performance.}
\begin{figure*}[t]
    \centering
    
    \begin{subfigure}[b]{0.59\textwidth}
        \centering
        \includegraphics[width=0.49\textwidth, height=0.183\textheight]{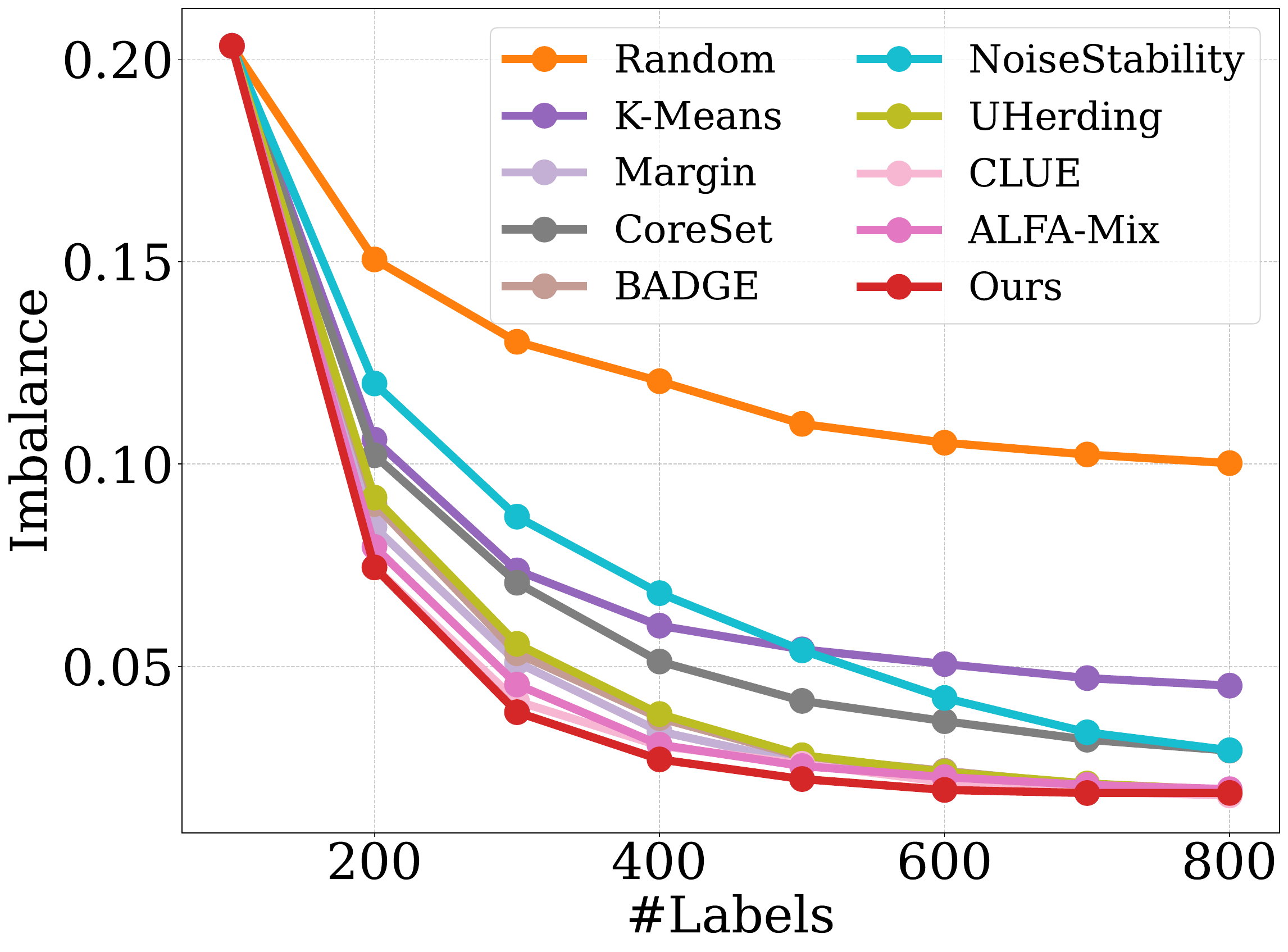}
        \hfill
        \includegraphics[width=0.49\textwidth, height=0.183\textheight]{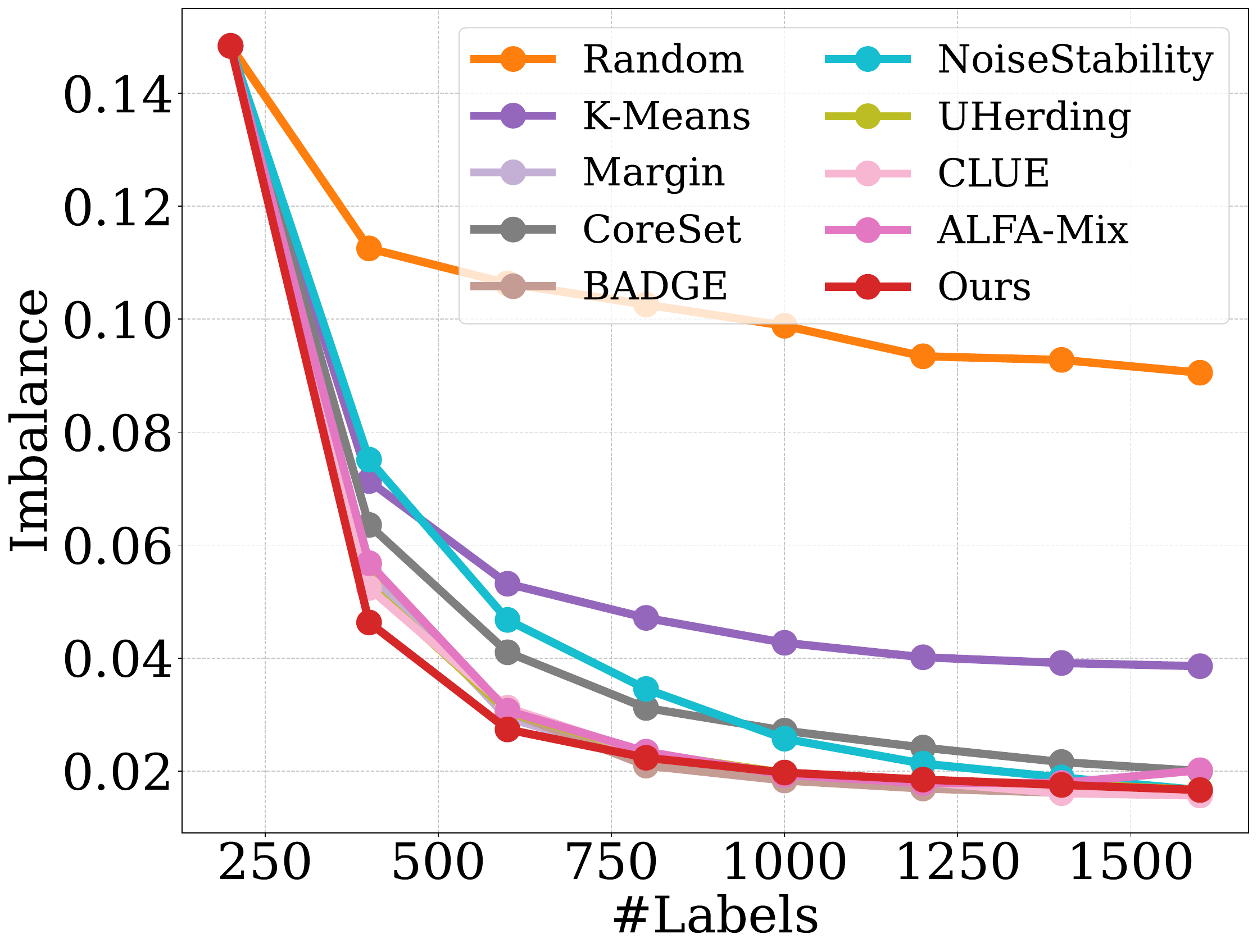}
        \caption{Imbalance of sampling data}
        \label{fig:R0107_imbl_entropy}
    \end{subfigure}
    \hfill
    \begin{subfigure}[b]{0.39\textwidth}
        \centering
        \includegraphics[width=\textwidth, height=0.178\textheight]{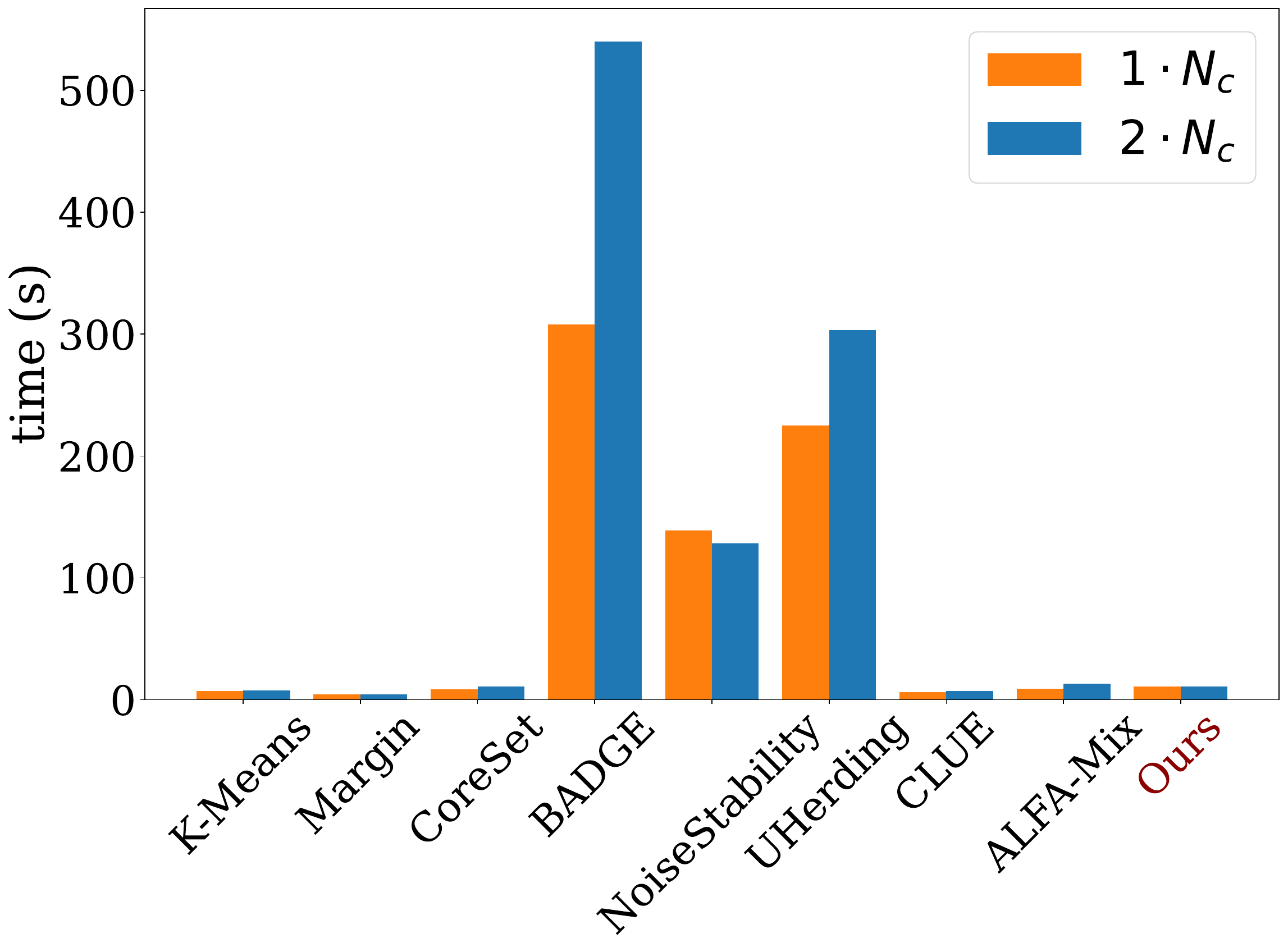}
        \caption{Data selection efficiency}
        \label{fig:time}
    \end{subfigure}

    \caption{
        (a) Imbalance of the sampling data by various AL methods through entropy.
        Each AL cycle we use the ResNet50 model to select \(K\cdot N_c\) samples in the Caltech101 dataset.
        \(N_c\) is the number of categories, \(K\in \{1,2\}\). 
        (b) Data selection efficiency of different methods.
        We compared the time required to select \(B=K\cdot N_c\) samples from Caltech101 dataset using the ResNet50 model per cycle.
    }
\end{figure*}
As illustrated in \cref{fig:R0107_imbl_entropy}, we quantify the average imbalance of the sampling data by various AL strategies using the class distribution entropy (see Appendix B.1).
Our approach achieves a well-balanced sampling data, thereby enhancing model performance.
In contrast to some methods, \eg, K-Means, CoreSet, and NoiseStability, which often induce performance degradation through imbalanced sample selection strategies.
This observation reveals that pronounced data imbalances significantly inhibit the potential performance gains of uncertainty-based or diversity-based sampling methods, particularly when such imbalances exceed critical thresholds.  

\textbf{Data selection efficiency.}
\begin{figure*}[t]
    \centering

    \begin{subfigure}[b]{0.19\textwidth}
        \centering
        \includegraphics[
            width=\textwidth,
            height=0.134\textheight,
        ]{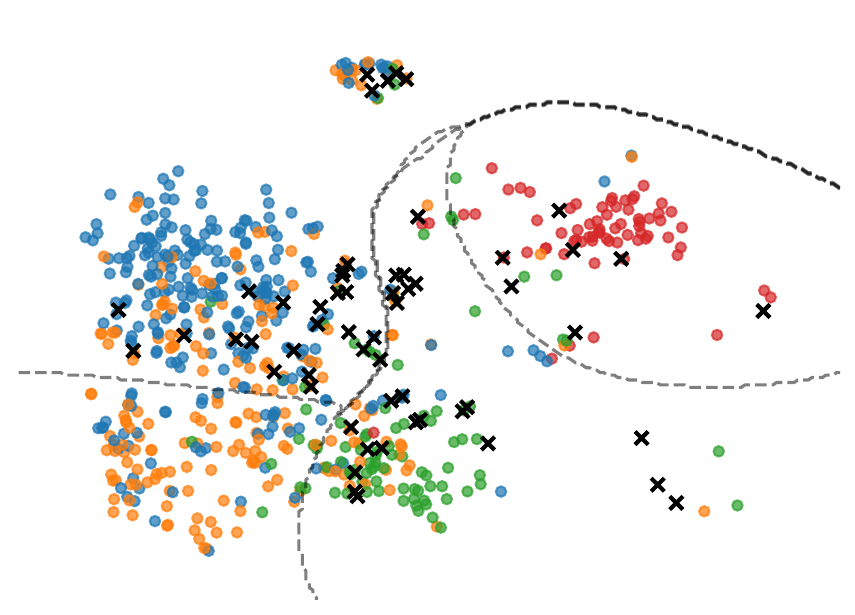}
        \caption{Ours}
    \end{subfigure}
    \hfill
    \begin{subfigure}[b]{0.19\textwidth}
        \centering
        \includegraphics[
            width=\textwidth, 
            height=0.134\textheight,
        ]{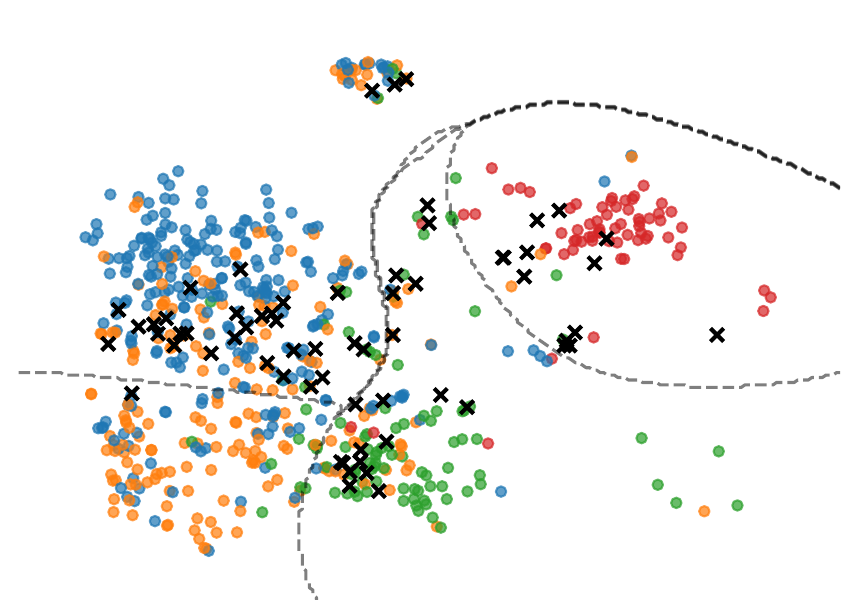}
        \caption{ALFA-Mix}
    \end{subfigure}
    \hfill
    \begin{subfigure}[b]{0.19\textwidth}
        \centering
        \includegraphics[
            width=\textwidth, 
            height=0.134\textheight,
        ]{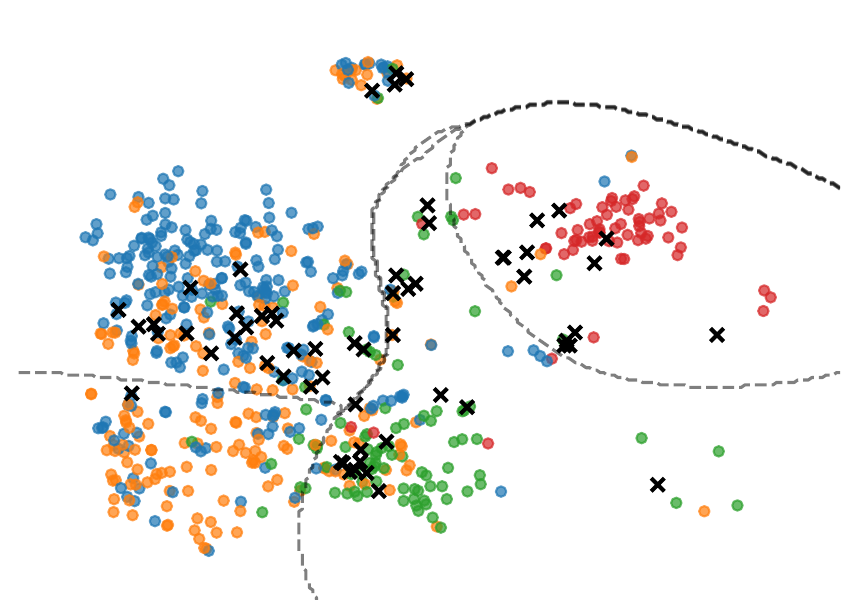}
        \caption{CLUE}
    \end{subfigure}
    \hfill
    \begin{subfigure}[b]{0.19\textwidth}
        \centering
        \includegraphics[
            width=\textwidth,
            height=0.134\textheight,
        ]{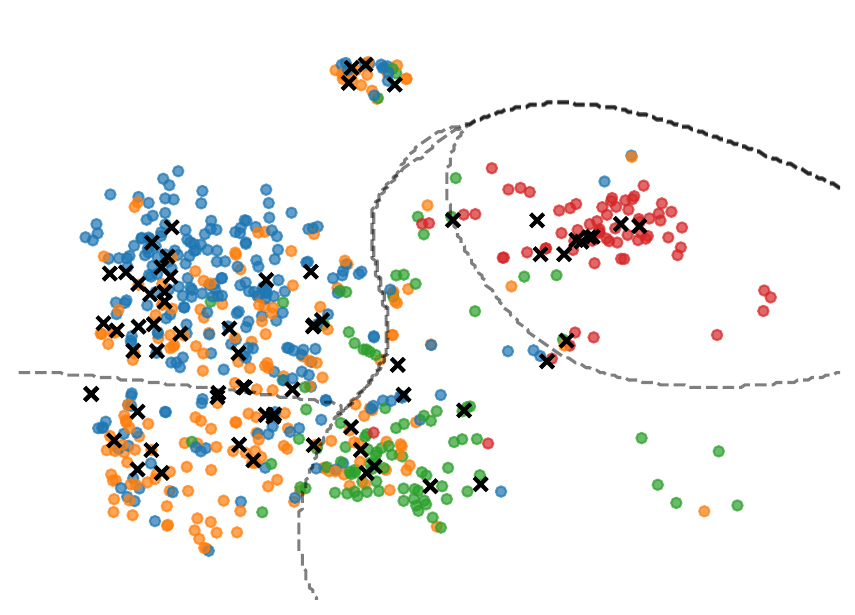}
        \caption{UHerding}
    \end{subfigure}
    \hfill
    \begin{subfigure}[b]{0.19\textwidth}
        \centering
        \includegraphics[
            width=\textwidth, 
            height=0.134\textheight,
        ]{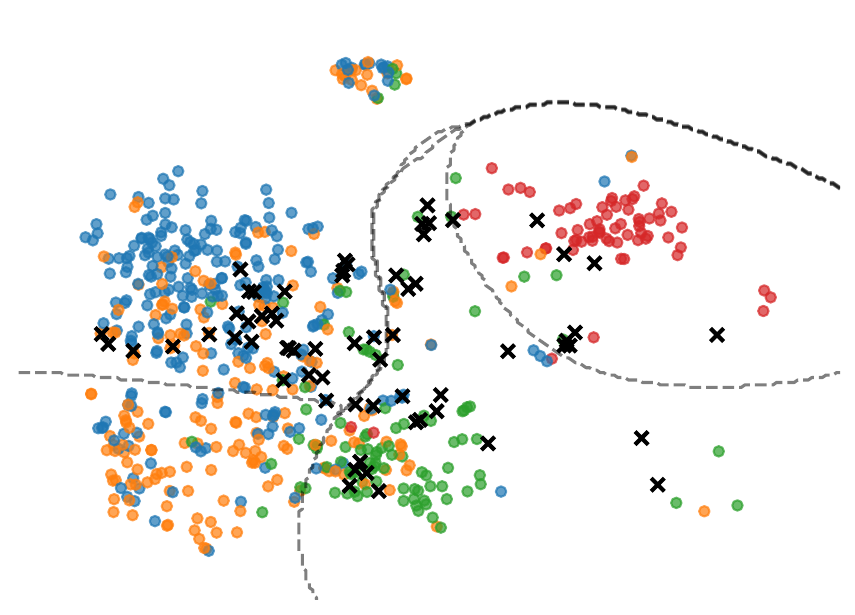}
        \caption{NoiseStability}
    \end{subfigure}
        
    \begin{subfigure}[b]{0.19\textwidth}
        \centering
        \includegraphics[
            width=\textwidth, 
            height=0.134\textheight,
        ]{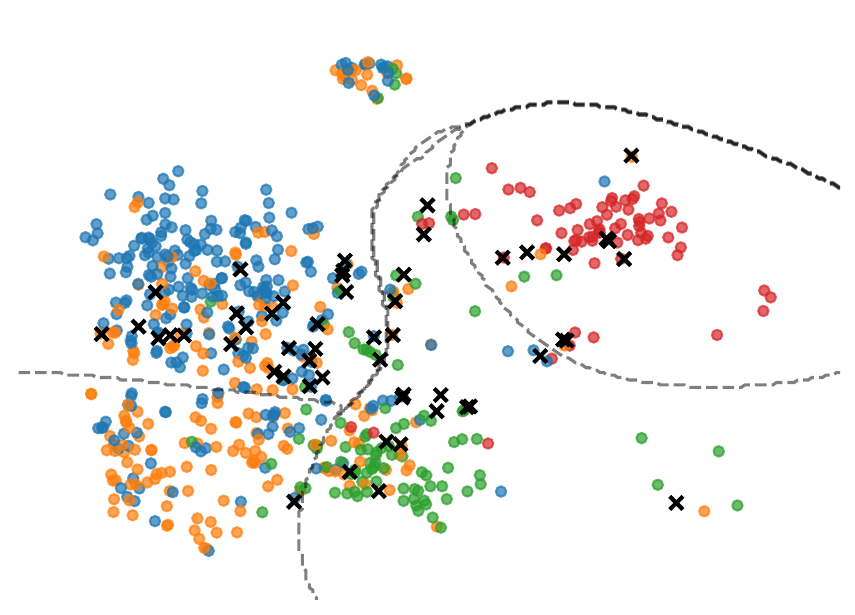}
        \caption{BADGE}
    \end{subfigure}
    \hfill
    \begin{subfigure}[b]{0.19\textwidth}
        \centering
        \includegraphics[
            width=\textwidth, 
            height=0.134\textheight,
        ]{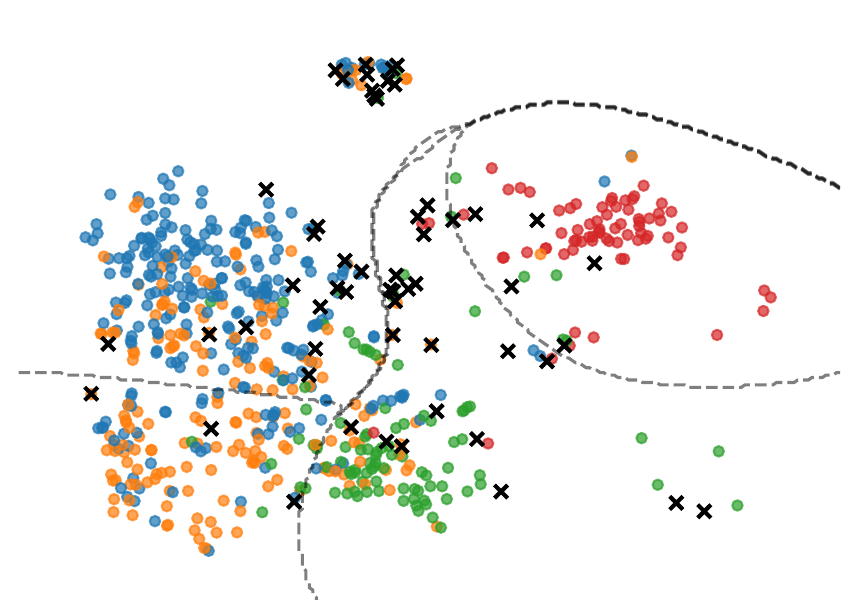}
        \caption{CoreSet}
    \end{subfigure}
    \hfill
    \begin{subfigure}[b]{0.19\textwidth}
        \centering
        \includegraphics[
            width=\textwidth, 
            height=0.134\textheight,
        ]{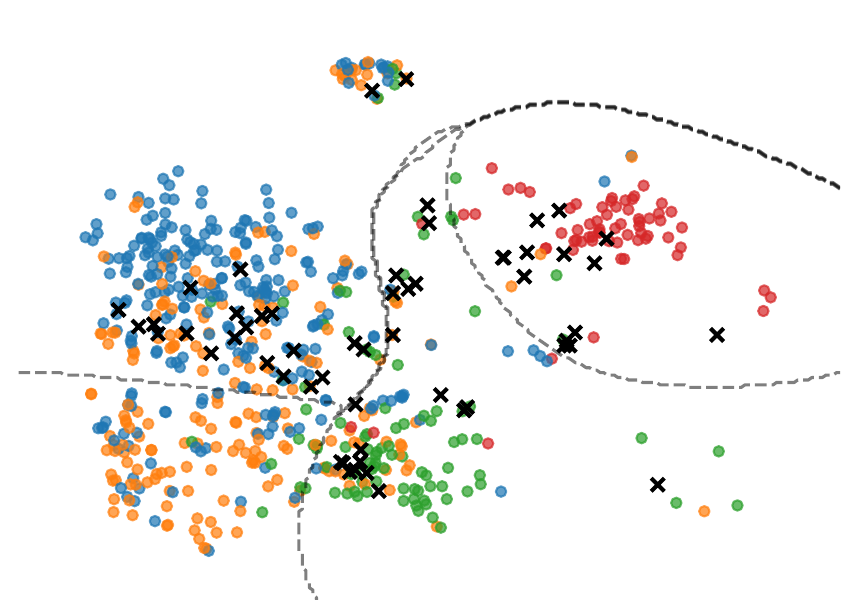}
        \caption{Margin}
    \end{subfigure}
    \hfill
    \begin{subfigure}[b]{0.19\textwidth}
        \centering
        \includegraphics[
            width=\textwidth, 
            height=0.134\textheight,
        ]{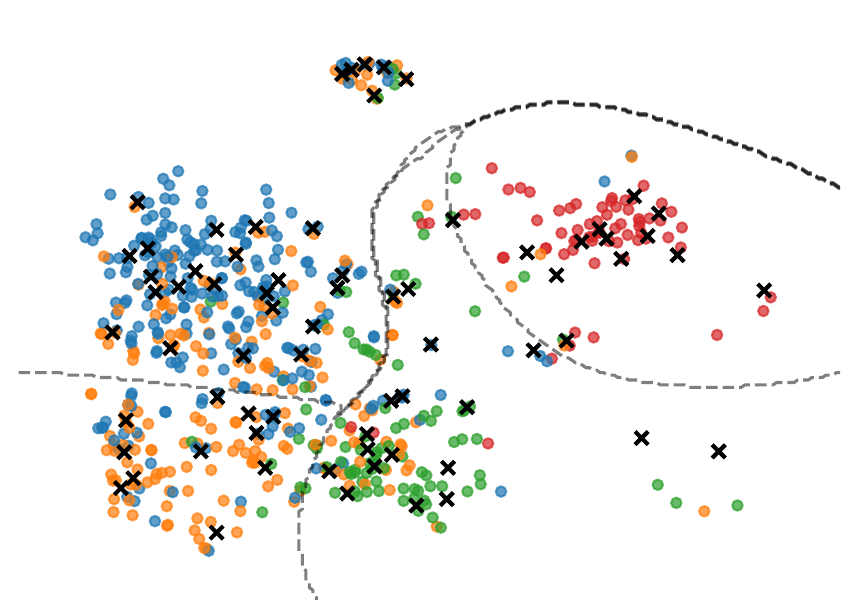}
        \caption{K-Means}
    \end{subfigure}
    \hfill
    \begin{subfigure}[b]{0.19\textwidth}
        \centering
        \includegraphics[
            width=\textwidth, 
            height=0.134\textheight,
        ]{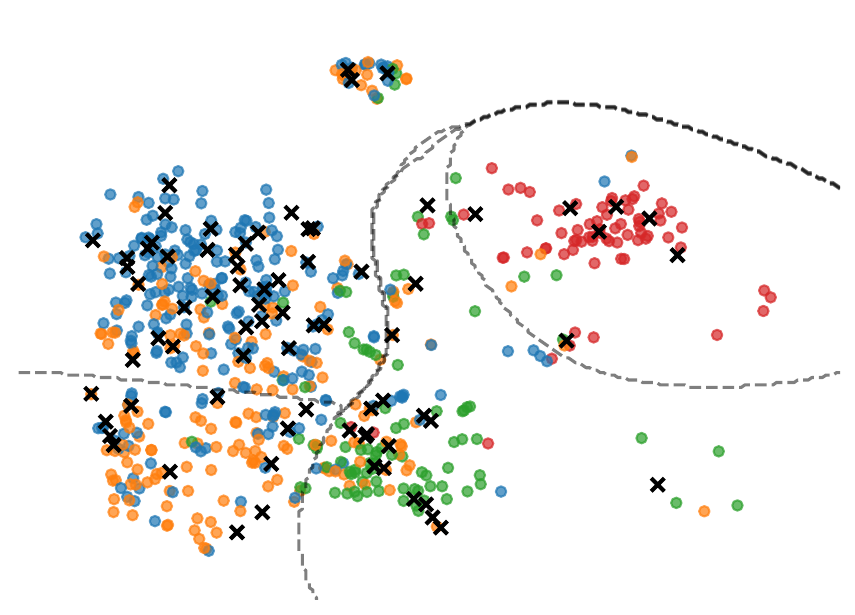}
        \caption{Random}
    \end{subfigure}
    
    \caption{
    t-SNE visualization of sample selection behavior by various AL methods on BronzeDing dataset with a ResNet50 model.
    The different colored dots represent 4 different categories of samples.
    Black forks indicate samples selected by various AL methods from the unlabeled pool, following an initial random labeling of 500 samples (with 100 additional samples sampled for annotation). 
    The oracle decision boundary is shown as a black dashed line.
    }
    
    \label{fig:tSNE}
\end{figure*}
We benchmark the time efficiency of selecting \(B\) samples per AL cycle on the Caltech101 dataset using the ResNet50 model, where the AL annotation budget \(B=K\cdot N_c\) is defined by the number of categories \(N_c\) and \(K\in \{1,2\}\). 
\cref{fig:time} shows that our method achieves a near-optimal time efficiency compared to the fastest baselines, while delivering substantially superior performance.
Furthermore, our method demonstrates a significantly lower data sampling time requirement than the BADGE, NoiseStability, and UHerding methods.
See Appendix C for additional discussion.

\textbf{Visualization of feature representations.}
To comparatively analyze sample selection behaviors, we visualize feature representations of samples selected by each AL strategy in \cref{fig:tSNE}.
These feature representations, extracted from the penultimate layer of the ResNet50 model, are projected into the 2D space via t-SNE~\cite{exam_tsne}.
We focus on four categories of the BronzeDing dataset, with the oracle decision boundary (dashed black line) displayed to enable a clear comparative analysis.
The results demonstrate that our proposed approach, while maintaining data diversity, effectively identifies and incorporates uncertain samples proximal to decision boundaries, thus enhancing the efficiency of active learning.

\subsection{Ablation Study}
\label{sec:ablation}
We implement ablation studies to validate the effectiveness of our design choices.
All ablation studies are conducted on Caltech101 datasets using the ResNet50 model with different AL annotation budget \(B\).
More details are provided in Appendix E.

\begin{wrapfigure}{r}{0.47\textwidth}
    \centering
    \captionof{table}{
        Ablation study of different components in our method.
    }
    \label{tab:component_ablation}
    \resizebox{0.47\columnwidth}{!}{
        \begin{tabular}{c|c|cc}
            \toprule
            \multirow{2}{*}{{Uncertainty}}        & \multirow{2}{*}{{Diversity}}                & \multicolumn{2}{c}{Caltech101, ResNet50} \\
            \hhline{~~--}
                                                  &                                             & \(B=1\cdot N_c\) & \(B=2\cdot N_c\) \\
            \midrule
            \textcolor[RGB]{255,99,71}{\ding{55}} & \textcolor[RGB]{255,99,71}{\ding{55}}       & 74.14\(\pm\)0.93 & 82.93\(\pm\)0.47 \\
            
            \midrule
            \textcolor[RGB]{255,99,71}{\ding{55}} & \textcolor[RGB]{50,205,50}{\ding{51}}       & 80.73\(\pm\)0.69 & 87.08\(\pm\)0.22 \\
            w/o \(\mathcal{S}_c\)                 & \textcolor[RGB]{50,205,50}{\ding{51}}       & 82.10\(\pm\)0.91 & 88.25\(\pm\)0.42 \\
            w/o \(\mathcal{S}_d\)                 & \textcolor[RGB]{50,205,50}{\ding{51}}       & 82.43\(\pm\)0.44 & 88.16\(\pm\)0.42 \\
            w \(\mathcal{S}_c\) & \textcolor[RGB]{255,99,71}{\ding{55}}       & 80.52\(\pm\)0.94 & 88.13\(\pm\)0.18 \\
            w \(\mathcal{S}_d\) & \textcolor[RGB]{255,99,71}{\ding{55}}       & 80.41\(\pm\)1.00 & 88.02\(\pm\)0.44 \\
            
            \midrule
            \textcolor[RGB]{50,205,50}{\ding{51}} & \textcolor[RGB]{255,99,71}{\ding{55}}       & 81.53\(\pm\)0.92 & 88.18\(\pm\)0.43 \\
            \textcolor[RGB]{50,205,50}{\ding{51}} & w/o Weighted                                & 82.39\(\pm\)0.42 & 88.18\(\pm\)0.59 \\
            \textcolor[RGB]{50,205,50}{\ding{51}} & w/o Clustering                              & 81.27\(\pm\)0.36 & 88.43\(\pm\)0.34 \\
            \textcolor[RGB]{50,205,50}{\ding{51}} & w/o Calibration                             & 82.94\(\pm\)0.88 & 88.04\(\pm\)0.42 \\
            
            \midrule
            \textcolor[RGB]{50,205,50}{\ding{51}} & \textcolor[RGB]{50,205,50}{\ding{51}}       & \textbf{82.95\(\pm\)0.37} & \textbf{88.49\(\pm\)0.37} \\
            \bottomrule
        \end{tabular}
    }
\end{wrapfigure}

\textbf{Effect of different components.}
Conceptually, we quantify the uncertainty of the data using both the discrepancy uncertainty \(\mathcal{S}_d\) and the confusion uncertainty \(\mathcal{S}_c\).
Ablation studies in \cref{tab:component_ablation} demonstrate that the removal of uncertainty sampling consistently induces performance degradation.
Moreover, reliance on a single uncertainty measure leads to suboptimal results, indicating that unidimensional uncertainty quantification fails to capture the nuanced heterogeneity inherent in fine-grained image data. 
Crucially, the absence of discrepancy uncertainty or confusion uncertainty prevents the model from learning features with poor structural stability and low category directionality. 
These unlearned features accumulate over AL cycles, progressively amplifying knowledge gaps, resulting in performance loss.

Our method selects diversity samples through calibration diversity after the uncertainty-weighted clustering.
Therefore, we ablate our diversity sampling strategy by four modified variants in \cref{tab:component_ablation}.
In general, our approach demonstrates consistent superiority over comparative variants, indicating the importance of calibration diversity.
We infer that over-reliance on local representativeness traps model in local optima, harming generalization, while prioritizing global diversity alone results in inefficient exploration and poor sample selection, hindering accuracy gains.

\textbf{Hyperparameter influence.}
\begin{figure*}[t]
    \centering
    \begin{subfigure}[b]{0.2425\textwidth}
        \centering
        \includegraphics[
            width=\textwidth, 
            height=0.151\textheight,
        ]{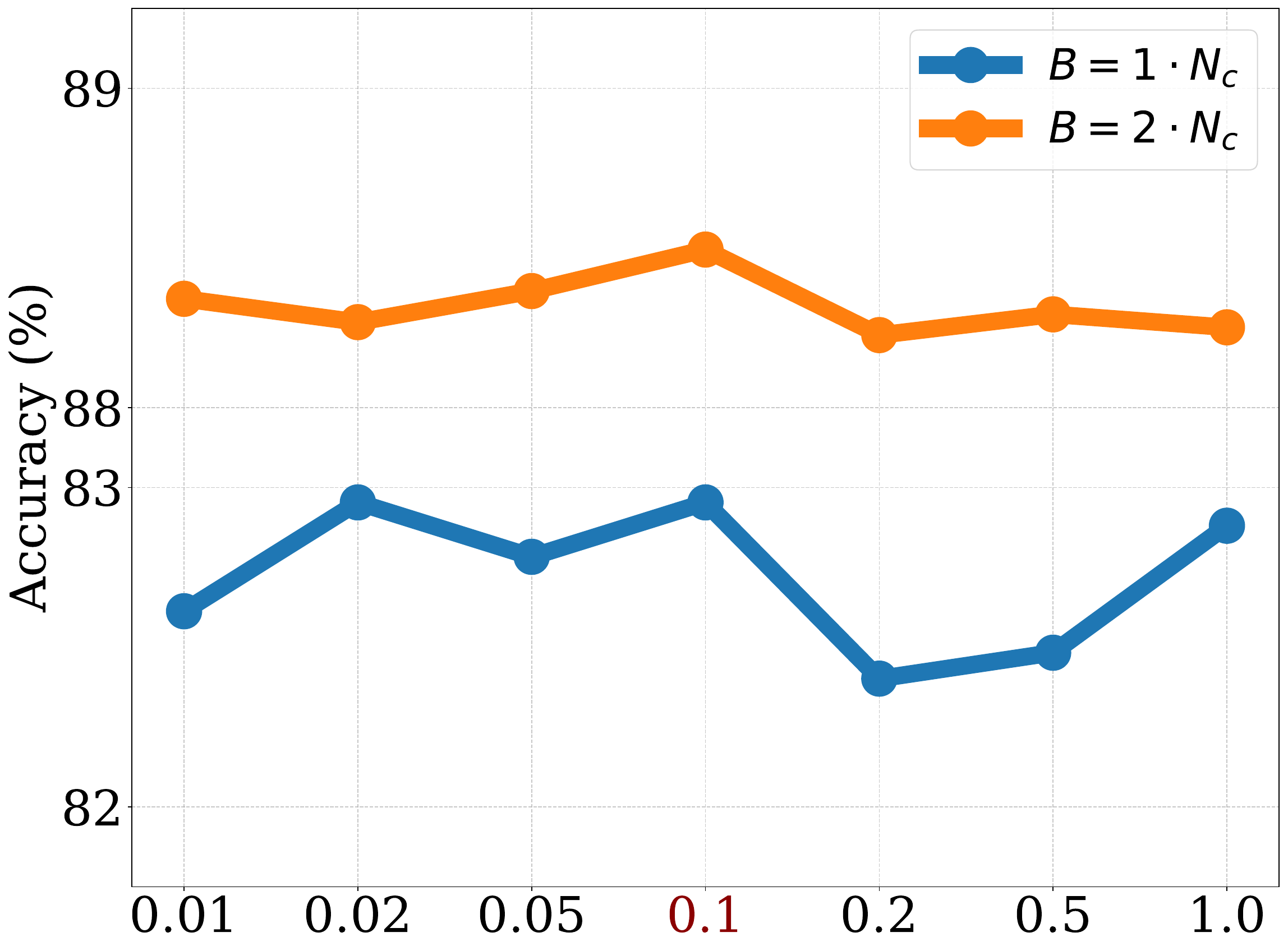}
        \caption{Effect of \(\eta\)}
        \label{fig:R0107_R0207_ps_MaskTopk_Acc7_Setup}
    \end{subfigure}
    \hfill
    \begin{subfigure}[b]{0.2425\textwidth}
        \centering
        \includegraphics[
            width=\textwidth, 
            height=0.151\textheight,
        ]{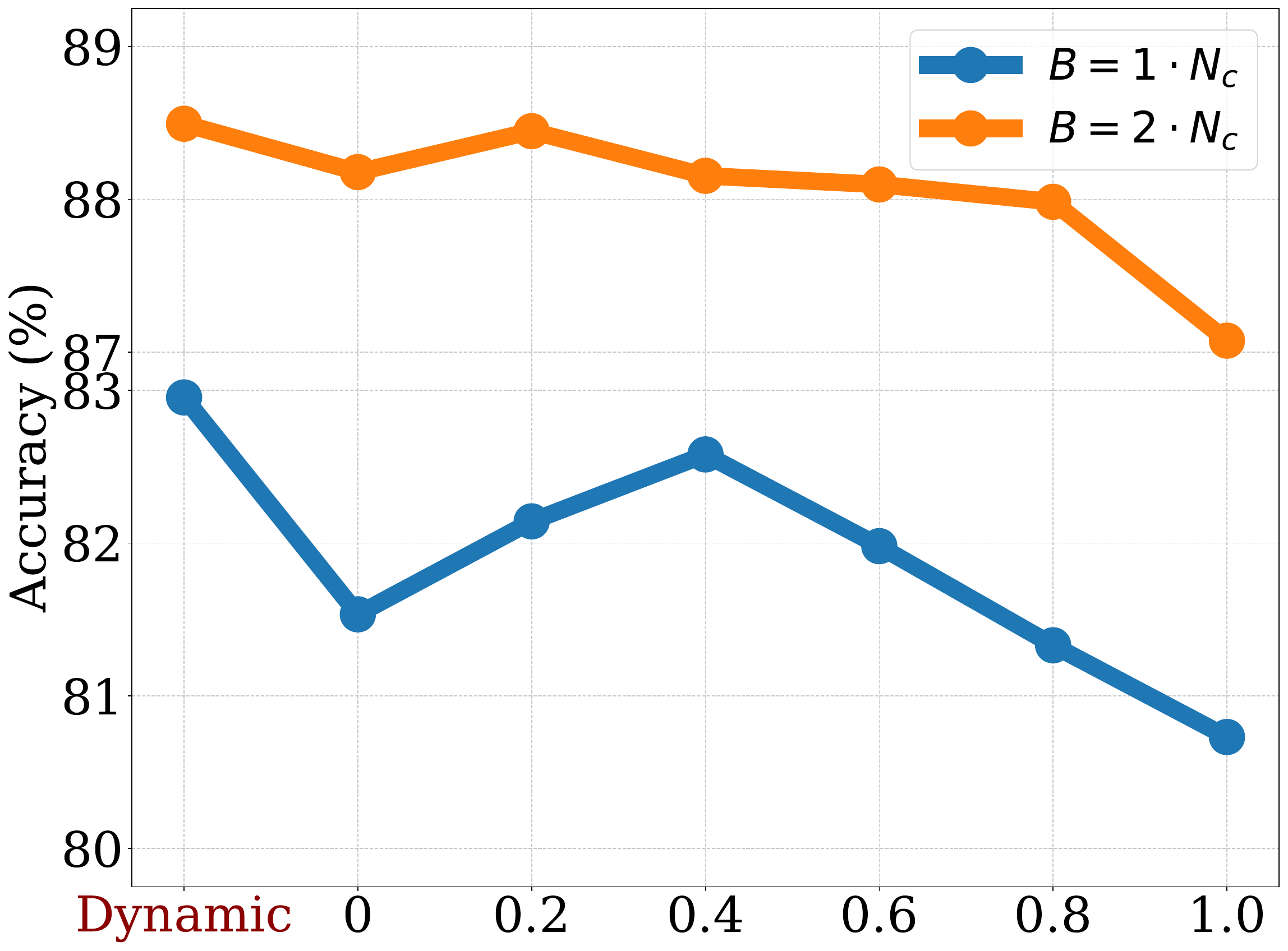}
        \caption{Effect of \(\zeta\)}
        \label{fig:R0107_R0207_ps_StaticGamma_Acc7_Setup}
    \end{subfigure}
    \hfill
    \begin{subfigure}[b]{0.2425\textwidth}
        \centering
        \includegraphics[
            width=\textwidth, 
            height=0.151\textheight,
        ]{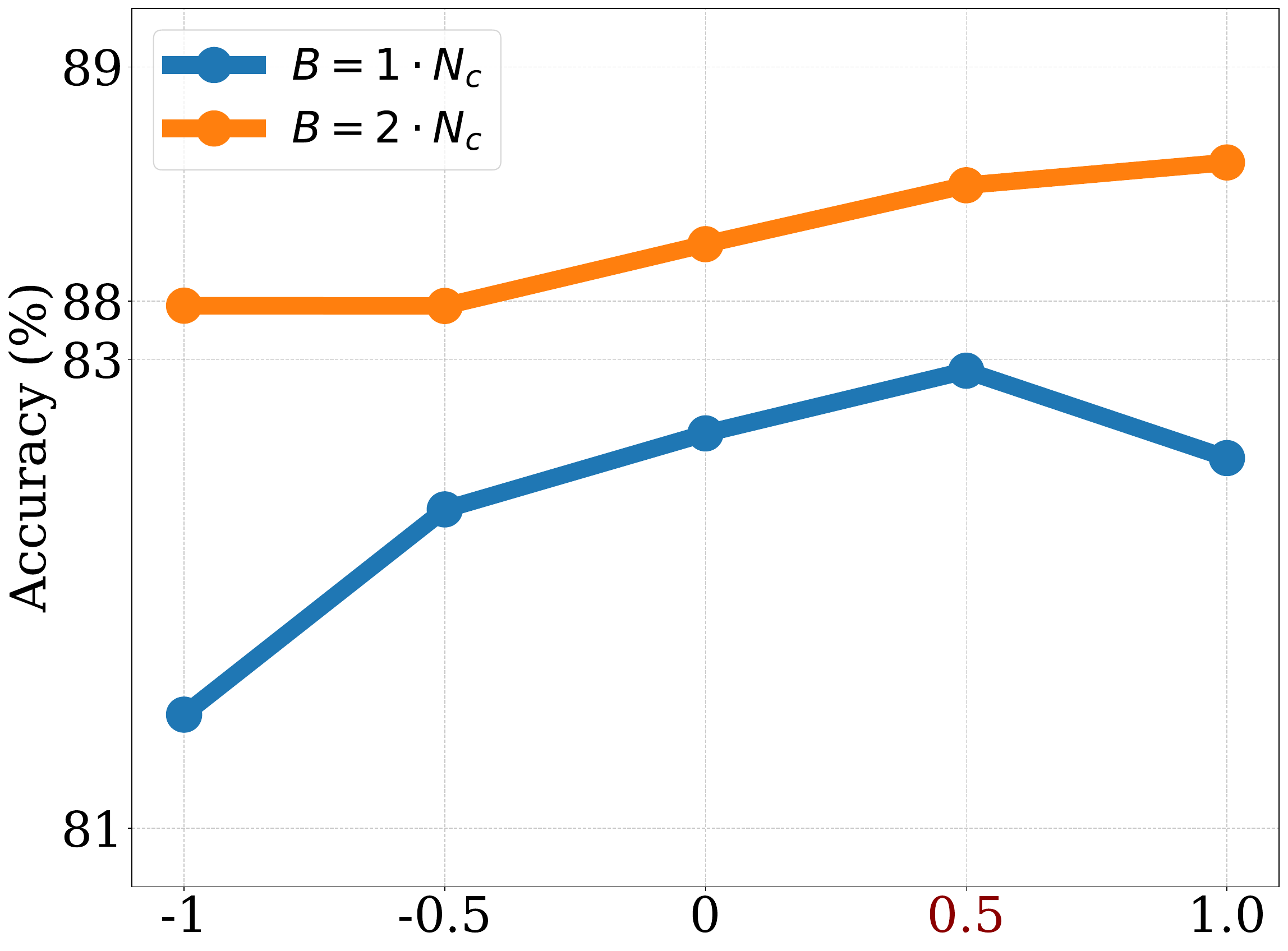}
        \caption{Effect of \(\lambda\)}
        \label{fig:R0107_R0207_ps_UncertaintyGamma_Acc7_Setup}
    \end{subfigure}
    \hfill
    \begin{subfigure}[b]{0.2425\textwidth}
        \centering
        \includegraphics[
            width=\textwidth, 
            height=0.151\textheight,
        ]{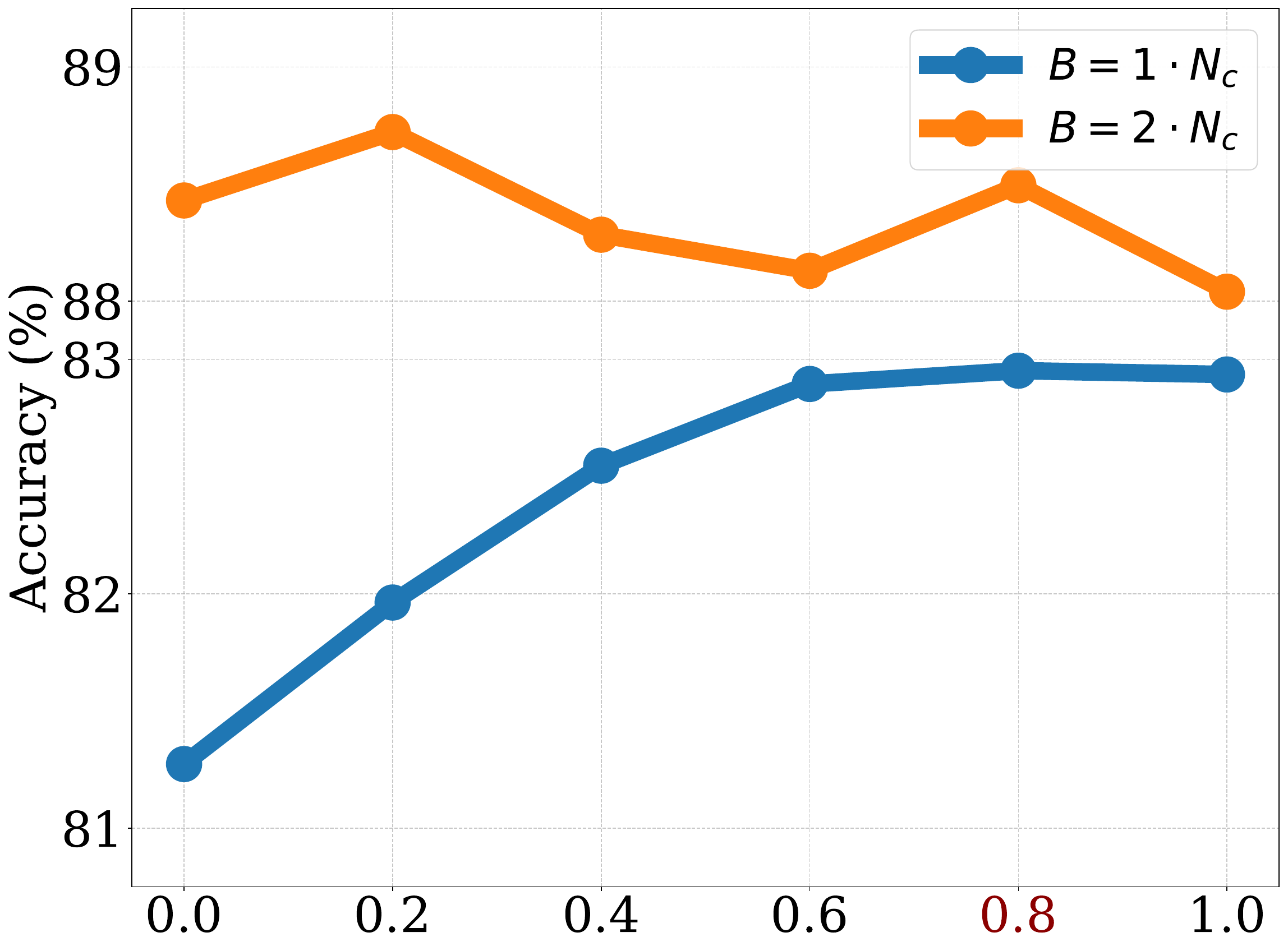}
        \caption{Effect of \(\xi\)}
        \label{fig:R0107_R0207_ps_DiversityGamma_Acc7_Setup}
    \end{subfigure}
    
    \caption{
        Ablations of the hyperparameters in our method.
    }
    \label{fig:ps_MaskTopk_StaticR_DiversityGamma}
\end{figure*}
In \cref{fig:ps_MaskTopk_StaticR_DiversityGamma},
we report four hyperparameter ablation studies,
including the local feature fusion size \(\eta\) in \cref{eq:mask},
the strength \(\zeta\) and the strength factor \(\lambda\) of uncertainty sampling in \cref{eq:threshold},
and the diversity balance factor \(\xi\) in \cref{eq:diversity_sampling}.

As shown in \cref{fig:R0107_R0207_ps_MaskTopk_Acc7_Setup}, the optimal performance is achieved with a moderate coefficient \(\eta=0.1\).
Insufficient local feature fusion (\ie, \(\eta\leq 0.05\)) fails to capture dominant patterns, preventing the AL strategy from effectively distinguishing probability offsets in unlabeled data post-fusion.
In contrast, a larger local fusion size (\ie, \(\eta\geq 0.2\)) introduces interference from contextual information, which affects the evaluation of sample values and effective selection.

For comparison, we set \(\zeta\) to fixed values and then select the percentage \(\zeta\) of the unlabeled data as candidates for diversity sampling.
As \cref{fig:R0107_R0207_ps_StaticGamma_Acc7_Setup} indicates, the fixed uncertainty sampling ratio does not accommodate active learning.
For a large fixed \(\zeta\), the candidates contain a large number of confidence samples, reducing the informativeness of the final selected sample and failing to refine the decision boundary.
In contract, uncertainty prevails when \(\zeta\) is small, which restricts the diversity of the selected data.
This leads to fragile distributions that struggle to perform consistently well when settings change.
The dynamic threshold \(\zeta\) in our algorithm that varies with the AL cycles is more adaptable to the demands of sampling over constant periods and achieves a trade-off between uncertainty sampling and diversity sampling.

As \cref{fig:R0107_R0207_ps_UncertaintyGamma_Acc7_Setup} indicates, a larger \(\lambda\) improves the model performance, particularly when the annotation budget \(B\) increases.
We attribute this to the fact that a larger \(B\) allows the model to develop a more robust understanding of the data distribution with a larger labeled dataset.
In this scenario, strengthening the intensity of uncertainty-based sampling is more beneficial, as it directs the model to focus on learning ambiguous patterns, thus refining and expanding the decision boundary.
In contrast, when model learning is still insufficient, an overemphasis on one side can be detrimental, and performance drops.

As \cref{fig:R0107_R0207_ps_DiversityGamma_Acc7_Setup} indicates, moderate diversity calibration (\ie, \(\xi=0.8\)) improves the quality of labeled data constructed with AL cycles and yield better performance.
However, when \(\xi\) is relatively small (\ie, \(\xi \leq 0.4\)) and the budget is small, the performance decreases as samples without local representativeness are selected.

\section{Conclusions}
We propose a novel active learning method, DECERN,
to improve fine-grained image annotation efficiency under limited budgets by combining discrepancy-confusion uncertainty and calibration diversity.
Specifically, DECERN perceives samples that exhibit poor structural stability and low category directionality during local feature fusion through discrepancy-confusion uncertainty,
and evaluates calibration diversity, considering both local feature representation and global diversity.
Through extensive evaluation on 7 fine-grained image datasets across 39 distinct experimental settings, our approach exhibits superior performance over state-of-the-art methods.

Although DECERN demonstrates efficient construction of high-quality labeled data under limited annotation budgets,
yielding substantial performance gains in fine-grained image classification,
its efficacy remains unproven when confronted by more complex tasks,
\eg, semantic segmentation, and object detection.
In future work, we plan to optimize our DECERN and implement our DECERN across more application scenarios.

\section*{Acknowledgements}
This work was supported by the National Natural Science Foundation of China (Grant No. 62576148).

%

\bibliographystyle{splncs04}
\bibliography{main}

\end{document}